\documentclass{article}
\usepackage[utf8]{inputenc}
\usepackage{authblk}
\usepackage{setspace}
\usepackage[margin=1.25in]{geometry}
\usepackage{graphicx}
\graphicspath{ {./figures/} }
\usepackage[labelfont=bf]{caption}
\usepackage{subcaption}
\usepackage{amsmath}
\usepackage{lineno}
\usepackage{xcolor}
\usepackage{soul}
\usepackage[normalem]{ulem}
\usepackage{amssymb} 
\usepackage{mathrsfs} 
\usepackage{longtable}
\usepackage{lscape} 
\usepackage{nicefrac}
\usepackage{fancyhdr}

\newcommand{\beginsupplement}{
    \setcounter{table}{0}
    \renewcommand{\thetable}{S\arabic{table}}
    \setcounter{figure}{0}
    \renewcommand{\thefigure}{S\arabic{figure}}
}

\usepackage[style=ieee, citestyle=numeric-comp, sorting=none]{biblatex}
\title{Few-Shot Learning Enables Population-Scale Analysis \\ of Leaf Traits in \textit{Populus trichocarpa}}

\author[1*]{John Lagergren}  
\author[1]{Mirko Pavicic}    
\author[1]{Hari B. Chhetri}  
\author[1]{Larry M. York}    
\author[1]{P. Doug Hyatt}    
\author[1]{David Kainer}     
\author[2]{Erica M. Rutter}  
\author[3]{Kevin Flores}     
\author[4]{Jack Bailey-Bale} 
\author[4]{Marie Klein}      
\author[4]{Gail Taylor}      
\author[1*]{Daniel Jacobson} 
\author[1*]{Jared Streich}   

\affil[1]{Biosciences Division, Oak Ridge National Laboratory, Oak Ridge, TN, USA.}
\affil[2]{Department of Applied Mathematics, University of California, Merced, CA, USA.}
\affil[3]{Department of Mathematics, North Carolina State University, Raleigh, NC, USA.}
\affil[4]{Department of Plant Sciences, University of California, Davis, CA, USA.}
\affil[*]{\texttt{\{lagergrenjh,jacobsonda,streichjc\}@ornl.gov}}

\date{} 

\begin{document}

\maketitle

\vspace{-3em}
\begin{abstract}
    Plant phenotyping is typically a time-consuming and expensive endeavor, requiring large groups of researchers to meticulously measure biologically relevant plant traits, and is the main bottleneck in understanding plant adaptation and the genetic architecture underlying complex traits at population scale. In this work, we address these challenges by leveraging few-shot learning with convolutional neural networks (CNNs) to segment the leaf body and visible venation of 2,906 \textit{P. trichocarpa} leaf images obtained in the field. In contrast to previous methods, our approach (i) does not require experimental or image pre-processing, (ii) uses the raw RGB images at full resolution, and (iii) requires very few samples for training (e.g., just eight images for vein segmentation). Traits relating to leaf morphology and vein topology are extracted from the resulting segmentations using traditional open-source image-processing tools, validated using real-world physical measurements, and used to conduct a genome-wide association study to identify genes controlling the traits. In this way, the current work is designed to provide the plant phenotyping community with (i) methods for fast and accurate image-based feature extraction that require minimal training data, and (ii) a new population-scale data set, including 68 different leaf phenotypes, for domain scientists and machine learning researchers. All of the few-shot learning code, data, and results are made publicly available. 
\end{abstract}

{\vspace{0.5em}\footnotesize\noindent{Notice: This manuscript has been authored by UT-Battelle, LLC, under contract DE-AC05-00OR22725 with the US Department of Energy (DOE). The US government retains and the publisher, by accepting the article for publication, acknowledges that the US government retains a nonexclusive, paid-up, irrevocable, worldwide license to publish or reproduce the published form of this manuscript, or allow others to do so, for US government purposes. DOE will provide public access to these results of federally sponsored research in accordance with the DOE Public Access Plan (\url{http://energy.gov/downloads/doe-public-access-plan}).}}


\section{Introduction} \label{sec:introduction}

Image-based plant phenotyping is a method by which scientists use image data to characterize and categorize plants within and across species. This process typically involves the use of tools, instrumentation, and domain expertise to (i) measure information from individual or groups of samples in the greenhouse, field, and/or nature, (ii) be applied across scales, ranging from cell microscopy to satellite imagery, and (iii) allow researchers to extract complex morphological and topological features that would otherwise be impossible to measure by hand. One of the main challenges of image-based phenotyping is identification of the relevant biological structures (foreground) from the background. In some cases, imaging methods can be modified to highlight these objects, such as using back lights or relying on fluorescence of those objects, however in many situations the contrast between the relevant object and the background is low. When contrast is high, simple greyscale or color-based thresholding can be used, but in more complex color imagery, plant phenomics has focused on machine learning approaches. 

Deep learning has revolutionized computer vision as a powerful and efficient way to extract features from image-based data~\cite{krizhevsky2017imagenet, he2016deep, ronneberger2015u}. An important strength of this approach is that deep learning models can learn an invariance to heterogeneous background effects which allows them to generalize to new samples outside of the training set. However, such approaches can be laborious and expensive to adopt, because users must generally annotate hundreds or thousands of images to provide sufficient training data. In plant biology for example, to reliably associate plant traits with genes at population scale requires large amounts of observations that span hundreds or thousands of genotypes. Such associations provide deeper understanding of the genetic architectures and underlying mechanisms that govern complex processes that control the growth, acclimation, response, and composition of plants, with important implications for sustainable agriculture and bioenergy~\cite{taylor2019sustainable, grattapaglia2018quantitative}. Thus, there exists a need to develop methods for fast and accurate image-based plant phenotyping that alleviate the data annotation bottleneck. 


In contrast to traditional deep learning approaches, which can require {large amounts} of training samples to reach sufficient prediction accuracy~\cite{krizhevsky2017imagenet, he2016deep}, few-shot learning is an emerging subset of machine/deep learning that attempts to maximize predictive accuracy while using only a small number of labeled samples for training. Multiple approaches exist to solve this problem, including data augmentation, metric learning, external memory, and parameter optimization~\cite{yang_fewshot}. This work utilizes a combination of data augmentation (i.e., applying random spatial and color augmentations to images during training) and iterative algorithms, which have been previously demonstrated for biomedical image analysis, e.g., semantic segmentation of cells and retinal vasculature~\cite{rutter_tracing, rutter_combo, januszewski_floodfilling, lagergren_growing}. The goal of this study is to extend these methods to image-based plant phenotyping by leveraging convolutional neural networks (CNNs) to segment the body and visible vein architecture of poplar (\textit{Populus trichocarpa}) leaves from high-resolution scans obtained in the field. In particular, few-shot learning is utilized in this work because it divides a small number of large images into a large number of small image tiles. In this way, the complex task of whole-image segmentation is broken down into smaller easier decision rules, which enables accurate segmentation using very few labeled images for training. {Note that use of the term ``few-shot learning'' to describe the methods in this work is distinct from the typical definition, and here refers to maximizing predictive outcomes using very few labeled images following terminology from~\cite{yang_fewshot}.}

\textit{P. trichocarpa} (also called black cottonwood, western balsam-poplar, or California poplar) is a model system for studying the genetic architecture of complex traits and climate adaptation in woody plants. Spanning from central California, USA, to northern British Columbia, Canada, it harbors tremendous geographic, climatic, phenotypic, and genetic diversity. Further, \textit{P. trichocarpa} has a fully sequenced genome, genome annotation, abundant transcriptomes, resequencing, and phenotypic data. Importantly, rapid biomass growth, clonal propagation, and the ability to grow in marginal lands with low agricultural input make it an ideal crop for sustainable bioenergy applications~\cite{tuskan2006, garcia2006protease, geraldes2011snp, zhang2018genome, slavov2012genome, evans2014population, chhetri2019multitrait, chhetri2020genome, slavov2012genome}. As a result, research and commercial groups have invested heavily in the development of \textit{P. trichocarpa} as a high-impact species for forest products and biofuel production~\cite{jansson2007populus, rubin2008genomics, evans2014population, mckown2014genome}. To this end, leaves play a key role in biomass production since they are the primary organs responsible for sunlight absorption and carbon fixation, the food source of vascular plant systems. Further, vein architecture supports the mechanical structure of the leaf and governs the distribution of water and other nutrients, which has important implications for the physiology, biomechanics, and structure of a plant~\cite{sack2013venation}. Thus, capturing accurate leaf traits and relating them to the genetic components that control them may provide insights toward improved tree biomass and composition.

In plant phenotyping, segmentation of individual leaves and their venation has seen sparse attention. In general, existing approaches use (i) experimental methods to chemically clear the leaf lamina and stain the veins to highlight the venation against the background~\cite{buhler2015phenovein, xu2021automated}, (ii) image pre-processing by greyscaling, aggregating specific color channels, or spatial rescaling~\cite{katyal2012leaf, larese2012legume, buhler2015phenovein, salima2015leaf, xu2021automated}, (iii) global filters and morphological operations (e.g., Odd Gabor filters, Hessian matrices, vesselness filters, and region merging) to obtain binary segmentations~\cite{katyal2012leaf, larese2012legume, buhler2015phenovein, salima2015leaf, zhu2020fast}, (iv) ensembles of scales and models to make aggregate predictions~\cite{zhu2020fast, xu2021automated}, and (v) require hundreds of manually-annotated training samples to produce accurate segmentation models~\cite{xu2021automated}. However, these commonly encountered steps can bottleneck the scalability and accuracy of image-based plant phenotyping at population scale. For example, approach (i) adds additional experimental time, effort, materials, expenses, and hazards to data acquisition compared to capturing just raw images, (ii) destroys fine-grained image details across spatial and color dimensions, (iii) may be overly simplistic and generate large amounts of effort in segmentation post-processing, (iv) uses complex workflows which may be difficult to automate at scale, and (v) can be infeasible for smaller research groups with limited time and budgets. These challenges may help explain why leaf and vein segmentation has not received as much attention compared to crop- or field-level phenotyping for plant stress, shoot morphology, and plant/organ counting~\cite{jiang2020convolutional}.

This work presents two few-shot learning methods based on CNNs to segment the body and visible vein architecture of \textit{P. trichocarpa} leaves. Leaf segmentation is formulated as a tracing task, in which a CNN iteratively traces the boundary of a leaf to produce a single contiguous leaf segmentation. Previous studies have shown that alternative CNN-based segmentation methods (e.g., the fully-convolutional neural network, U-Net~\cite{ronneberger2015u}) {that do not include spatial priors that encourage object contiguity may result in biologically unrealistic segmentation masks (i.e., contain holes or other artifacts caused by individual or groups of pixels that do not exceed a global probability threshold)}~\cite{rutter_tracing}. In contrast, boundary tracing {addresses this challenge} by only segmenting one contiguous region, thereby ensuring accurate downstream extraction of morphological features. Alternatively, vein segmentation is formulated as a region growing task, in which a CNN iteratively adds neighboring pixels to a growing region of interest corresponding to the visible vein architecture. Similar to the tracing approach, the vein segmentation ensures biologically-realistic morphological features by including pixels in the segmentation only if a neighboring pixel was previously classified. Each method is fully automated (i.e., requires no human supervision or initialization), and segments images orders of magnitude faster compared to manual annotation. {Additionally, each method is compared against a variant of the state-of-the-art image segmentation model, U-Net~\cite{ronneberger2015u}.}

The current work is designed to provide the plant phenotyping community with (i) methods for fast and accurate image-based feature extraction with minimal training data and (ii) a new population-scale data set for domain scientists and machine learning researchers. In particular, the methods developed here are applied to raw RGB images with no experimental/image pre-processing, use individual CNN models that learn the complex relationships between pixels for accurate leaf and vein segmentation, and require very few training samples to generalize and make accurate predictions at population scale. The segmentations are used to extract biologically realistic features that are validated using real-world physical measurements and applied downstream using broad-sense clonal heritability estimates and a genome-wide association study (GWAS).

\section{Materials and Methods} \label{sec:methods}

CNNs are used to segment the body and visible vein architecture of \textit{P. trichocarpa} leaves from high-resolution scans. The resulting segmentations are combined with open-source tools for image processing and genomic analysis to expand the application of these methods to a wider scientific audience. All deep learning methods are implemented in Python (version 3.7.8) using the PyTorch deep learning library (version 1.11.0)~\cite{paszke2019pytorch} and {are made publicly available at~\cite{fsl_code}}. Feature extraction is completed using Fiji (version 2.9.0)~\cite{schindelin2012fiji} and RhizoVision Explorer (RVE, version 2.0.3)~\cite{seethepalli2020rhizovision, seethepalli2021rhizovision}. Genomic analysis is conducted in R (version 4.2.0) using the GAPIT3 software package (version 3)~\cite{wang2021gapit}. All of the images, manual segmentations, model predictions, extracted features, and underlying genome sequences {are made publicly available at~\cite{fsl_code, fsl_data}}.


\subsection{Data collection} \label{sec:data}

The leaf scans considered in this work were collected during a field campaign in August, 2021, from the 10-acre poplar plantation at the University of California, Davis (UC Davis), which maintains a common garden of poplar trees that can be grown on low-quality, poor, and marginal land~\cite{baileybale2021plantation, taylor2019sustainable}. The plantation {is} composed of three blocks. Each block is partitioned into rows and positions that uniquely identify the corresponding genotypes, and contain approximately 1,500 \textit{P. trichocarpa} trees per block. For practical reasons, leaf samples were collected from one entire block (1,322 viable samples) and partially from a second block (131 samples) totaling 1,453 trees. 

Leaves were sampled from a branch at approximately breast height (i.e., $\sim$1.37 meters) from the south-facing side of each tree. Leaves were chosen by selecting the first fully mature leaf counting from the top of each branch. Each leaf was also paired with a barcode label that encoded the treatment, block, row, and position of the tree, which uniquely identified the corresponding genotype and allowed the user to record the sample ID during data capture. This helped expedite the phenotyping process and reduce human error. Selected leaves were scanned in the field as they were sampled from each tree using a USB-powered Epson Perfection model V39~\cite{epsonwebsite}. The top and bottom of each leaf was scanned with a resolution of 300 dots-per-inch (DPI). To account for heterogeneous leaf shapes (e.g., leaves with non-trivial 3D characteristics like ``waviness''), a weight was used on the scanner lid to compress each leaf to the glass of the scanner in order to reduce image artifacts like blurring. Additionally, between rows of trees (there are approximately 30 trees per row), the scanner glass and background were cleaned to reduce the buildup of dust and other debris. 

During data capture, the scanner suffered a hardware failure where one of the pixels of the scanner began to malfunction and caused a vertical white line to gradually appear near the center of each subsequent scan. This artifact affected approximately 100 leaf scans. To mitigate the malfunction, leaves were moved to the edge of the scanner away from the malfunctioning pixel, affecting 62 leaf scans. A new scanner was acquired and used for the remainder of the field campaign (2,634 leaf scans). Despite the hardware failure, these data acquisition steps resulted in 2,906 RGB leaf scans (i.e., top and bottom of 1,453 samples), each with dimension $3510\times2550$ pixels. 

In addition to image-based measurements, petiole length and diameter were measured manually for each leaf. Using a similar procedure to leaf imaging, barcode scanners were used to record the sample ID, followed by length/diameter measurements using USB-powered SPI 17-600-8 electronic calipers~\cite{spiwebsite}. The manual measurements for petiole length and width are used to validate image-based measurements.

Obtaining accurate high-quality ground truth data is important for deep learning applications in general, but it is crucial for few-shot learning, since a model must learn features from a small number of training samples that generalize well to the broader population. To this end, training data was generated for leaf body segmentation using the top and bottom scans of 25 leaves (50 images in total), which were randomly selected and manually traced. Manual segmentation was completed using the open-source graphics editor, GNU Image Manipulation Program (GIMP)~\cite{gimp}, taking between 15 and 30 minutes per image, depending on the size and serration of the leaf. Similarly for vein segmentation, GIMP was used to manually draw all visible leaf venation for eight leaf-bottom scans, taking between four and eight hours per image, depending on the vein density. Note that only leaf-bottom venation is considered in this work. Leaf-top venation will be considered in future work.

Due to the large amount of manual effort required for vein segmentation, the training data set was constructed using \emph{iterative data set refinement}, in which images were individually added to the training set based on manual inspection of model performance across the set of all images. For example, compressing samples against the scanner glass caused some leaves to fold on themselves, which produced dark lines that were falsely identified as veins. Thus, an image with multiple examples of such folds was manually segmented and added to the training set so that the model learned an invariance to such artifacts. This process was repeated similarly for other leaf characteristics (e.g., dead, diseased, and nutrient-deficient leaf tissue), including a scan exhibiting the hardware failure discussed above, until the model converged to acceptable performance across the population. This strategy resulted in a total of eight images (six for training, two for validation) mentioned previously. Note that in practice, the number of images may vary depending on the application and image quality, but it is important (particularly for few-shot learning) that the training data set is fine-tuned to the point that the model is able to generalize.

\subsection{Leaf segmentation} \label{sec:tracing}

\begin{figure}[h!]
    \centering
    \includegraphics[width=\textwidth]{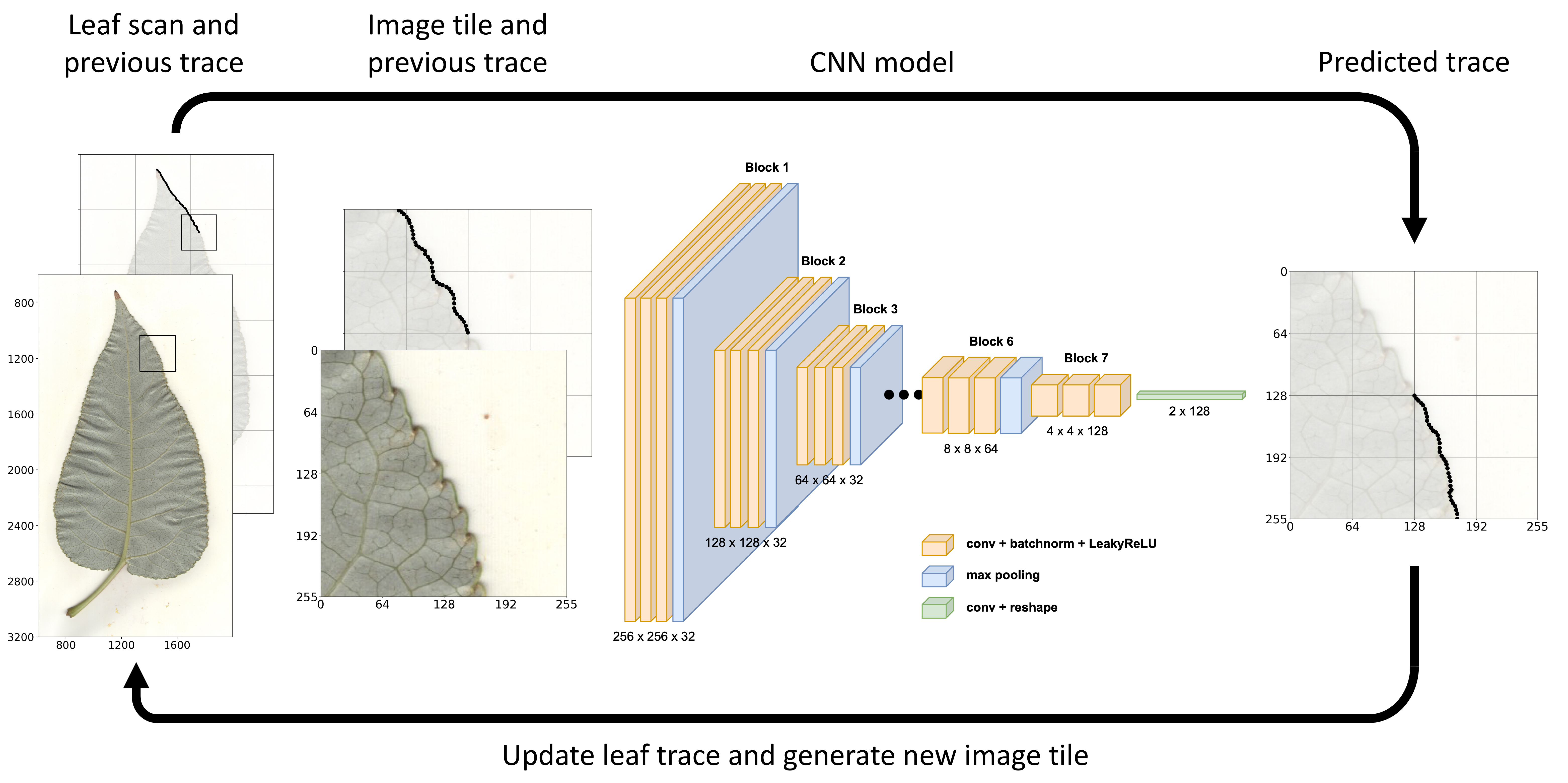}
    \caption{\textbf{Leaf tracing algorithm.} An image tile and a small segment of the previously traced path are input to a CNN which predicts the next steps of the trace. The predictions are added to the leaf contour and used to generate an image tile at the new location. This iteration continues until the trace reaches the starting location of the contour. \underline{Left}: RGB leaf scan and input tile (front) with previously traced pixels (back). \underline{Center}: the leaf tracing CNN which transforms the $256\times256$ input tile into a $2\times128$ set of pixel displacements for trace prediction. \underline{Right}: predicted pixel displacements that are used to update the trace and generate image tiles in the next iteration.}
    \label{fig:tracer}
\end{figure}

Segmentation of the leaf body is formulated as an object tracing task based on~\cite{rutter_tracing, rutter_combo}, in which a CNN is used to iteratively trace along the contour of a leaf. These methods have been shown to reach state-of-the-art accuracy in biomedical image segmentation using a fraction of the training data required by other approaches~\cite{rutter_combo}. In this framework, a CNN inputs a small image tile centered somewhere on the edge of an object and outputs a predicted trace (i.e., set of pixel displacements) along the object boundary from the center to the edge of the tile. The iteration proceeds by generating new image tiles along the predicted contours, continuing the trace until reaching the starting location, thereby closing the loop and finishing the segmentation. An important benefit of this approach is that it breaks the complex task of whole-leaf segmentation into multiple smaller, easier decision rules, and requires only a small number of images to train an accurate model. See Figure~\ref{fig:tracer} for a diagram of the leaf tracing algorithm and Supplementary Video S1 for a video of the iteration.

The leaf tracing CNN inputs $256\times256\times4$ image tiles, which include three color channels (RGB) and an overlay of the previously traced path as an additional channel. The tile size is chosen large enough to provide the model with sufficient context to trace through areas where the leaf contour may be obscured (e.g., in damaged/diseased areas or near the petiole). The RGB values are normalized to $[0, 1]$ for computational stability. The additional channel is a binary image comprised of ones along pixels of the previously traced path and zeros otherwise, and thus provides the network with a direction to continue the trace. Each image tile is centered at a pixel on the contour of a leaf, by which the 50 manually-traced samples are used to generate more than 300,000 individual tiles for training. Further, heavy image augmentation is used so that the CNN learns an invariance to heterogeneous leaf shapes and conditions. In particular, random continuous rotations, horizontal and vertical flips, displacement jitter, and color augmentation (hue, saturation, brightness, and contrast) are combined so that no two image tiles appear the same during training.

The leaf tracing CNN outputs $2\times N$ trace predictions, which encode $N$ horizontal and vertical pixel displacements along the leaf contour relative to the center pixel of the input tile. Training data is generated by evenly sampling pixels from the center to the edge of the tile along the contour of the leaf. Distance is then measured between the predicted trace and the ground truth contour using mean squared error, $\mathcal{L}_{\text{MSE}}$, as an objective function. Importantly, the quality of the predicted trace degrades near the edges of image tiles since the CNN does not have context beyond the boundaries of the input. However, it is still important for the model to predict the trace from the center pixel to the edge of the image tile so that the predicted trace can ``skip'' over obscured segments of the contour~\cite{rutter_combo}. To account for these effects, weighted mean squared error is used to weight predictions closer to the center pixel more heavily than predictions near the edge. The objective function is given by

\begin{subequations}
\begin{align}
    \mathcal{L}_{\text{MSE}} &= \frac{1}{N} \sum_{i=1}^{N} \omega_i \big\| y_i - \text{CNN}(x)_i \big\|_2^2,
    \label{eq:mse}
    \\
    \omega_i &= 1 + \frac{1 - \tanh\left(\alpha i + \beta\right)}{2},
    \label{eq:tanh}
\end{align}
\end{subequations}

\noindent where $x \in \mathbb{R}^{256\times256\times4}$ is the input image tile, $y \in \mathbb{R}^{2\times N}$ is the set of ground truth row/column coordinates with $y_i$ indicating the row and column position of the $i^{\text{th}}$ pixel, $\omega \in \mathbb{R}^{N}$ is the weight vector, the number of pixel displacements is $N=128$, and $\alpha = 8/N$ and $\beta = -4$ are chosen such that the hyperbolic tangent function (which defines $\omega$) gradually decreases the error weight from two to one along the predicted contour. In this way, the objective function weights pixels near the center of the tile approximately two times greater than pixels near the edge.

The model architecture follows standard practices for CNNs~\cite{simonyan2014very, he2016deep}. In particular, the CNN uses blocks of three $3\times3$ convolution layers with zero-padding and one max pooling layer. Each convolution layer includes batch normalization to stabilize training~\cite{ioffe2015batch} and a ``LeakyReLU'' activation function for nonlinearity~\cite{maas2013rectifier}. Additionally, residual connections are applied between convolution layers for easier optimization and better prediction accuracy~\cite{he2016deep}. In total, the leaf tracing CNN includes six blocks with max pooling and one block without, which transforms the spatial image dimensionality from $256\times256$ to $4\times4$. Then, a final $4\times4$ convolution layer reduces the outputs to a vector of length 256, which is reshaped into $2\times128$ for trace prediction. Note that the final convolution is linear (i.e., it does not include a nonlinear activation function) so that the trace predictions can reach the edges of the input tile in any direction. To prevent overfitting, images are randomly split into 80\% training (i.e., 40 images totaling $\sim$250K image tiles) and 20\% validation (i.e., 10 images totaling $\sim$50K image tiles) sets. The model is then trained for 1,000 epochs with a batch size of 256 and the Adam optimizer~\cite{kingma2014adam} with default parameters. Further, early stopping with 20 epochs (i.e., training is stopped if the validation error does not improve within 20 epochs) is used to guarantee the convergence of the model.

Once the leaf tracing CNN is trained, it is used to iteratively trace the contour of each leaf image in the data set. The tracing algorithm is initialized using automatic thresholding to obtain a rough segmentation of the leaf, which provides both a starting location and trace direction. An image tile centered at the top of the rough segmentation (i.e., at the tip of the leaf) is initially fed to the CNN, which outputs the initial trace prediction from the center to the edge of the image tile. The first 32 pixel predictions along the edge of the leaf are added to the trace, and a new image tile is drawn centered at the new location. This iteration continues until the predicted trace falls within 10 pixels of the previously-traced contour, after which a line is drawn from the prediction to the contour to close the loop. To eliminate errors from the trace initialization, the tracing algorithm uses 10 ``burn-in'' iterations before storing traced pixels for the final segmentation. The leaf body segmentation is obtained by classifying all interior pixels as foreground and exterior pixels as background. Note that the trace direction is randomized during training, so that the tracing CNN can segment leaves in either clockwise or counterclockwise directions. In practice, the tracing algorithm does not require human supervision to start or stop the iteration and takes {$\mathcal{O}(1)$} second per image on a single GPU of an NVIDIA DGX Station A100.

\subsection{Vein segmentation} \label{sec:growing}

\begin{figure}[h!]
    \centering
    \includegraphics[width=\textwidth]{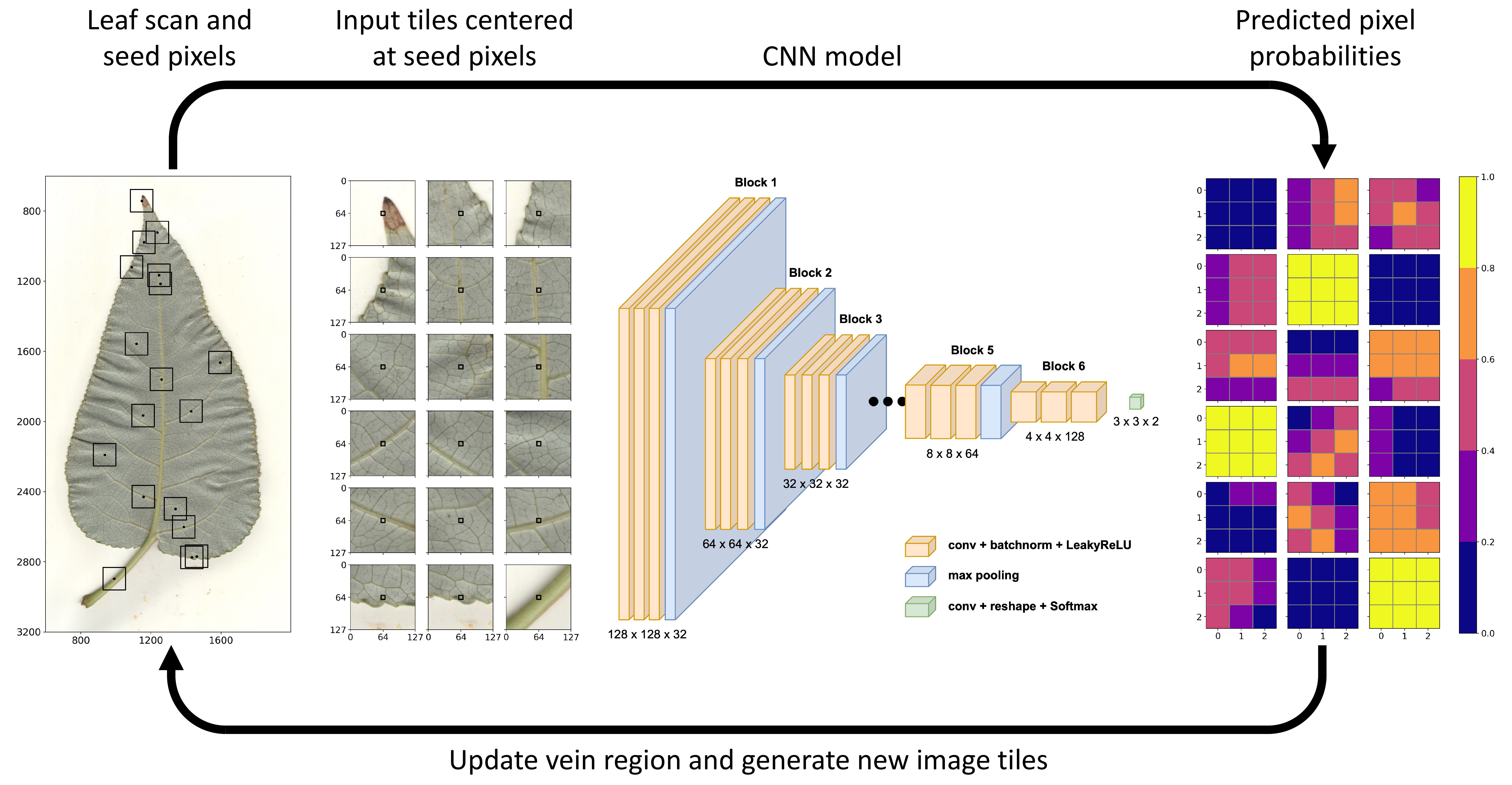}
    \caption{\textbf{Vein growing algorithm.} Image tiles centered on pixels of interest are input to a CNN which predicts the classification of the center pixel and its neighbors. Neighboring pixels with high probability are added to the vein region and used as seed pixels in the next iteration. The iteration continues until no new pixels are added to the vein region. \underline{Left}: RGB leaf scan and input tiles with center pixels highlighted in black. \underline{Center}: the vein growing CNN which transforms the $128\times128$ input tile into a $3\times3$ matrix of vein probabilities. \underline{Right}: predicted pixel probabilities that are used to update the region and generate new image tiles in the next iteration.}
    \label{fig:grower}
\end{figure}

Segmentation of the leaf venation is formulated as a region growing task based on~\cite{januszewski_floodfilling, lagergren_growing}, in which a CNN is used to iteratively expand a region of interest (i.e., visible veins of a leaf). The convolutional region growing method (also called flood filling networks~\cite{januszewski_floodfilling}) has been shown to reach state-of-the-art segmentation accuracy while preserving biologically realistic morphological features~\cite{lagergren_growing}. However, rather than tracing the boundary of an object with a 1D line, the vein growing CNN iteratively grows a segmentation in all directions (e.g., 2D in~\cite{lagergren_growing} and 3D in~\cite{januszewski_floodfilling}) by classifying which pixels/voxels should be included in or rejected from the region. In particular, a CNN inputs small image tiles centered on pixels of interest and predicts classifications of the center pixel and its adjacent neighbors. Neighboring pixels that are added to the region become the seeds for new image tiles in the next iteration. This process continues until no new pixels are added to the region, thereby finishing the segmentation. Similar to the leaf tracing framework, the region growing approach breaks the complex task of vein segmentation into many smaller decision rules and can produce high-accuracy segmentations using fewer than ten images for training. See Figure~\ref{fig:grower} for a diagram of the vein growing algorithm and Supplementary Video S2 for a video of the iteration.

The vein growing CNN inputs $128\times128\times3$ RGB image tiles (also normalized to $[0, 1]$) centered on pixels in the interior of a leaf. The tile size is chosen to be smaller than the leaf tracing tiles (i) since vein classification does not require as much context and (ii) for computational efficiency, since many more image tiles are used in this framework. However, the tile size is still large enough so that the model can accurately predict vein pixels in areas of uncertainty (e.g., blurry patches and diseased/dead tissue). To construct a training set, image tiles are drawn for each vein pixel, by which the eight manually-segmented images generate more than 1,000,000 positive samples (i.e., samples centered on leaf veins rather than leaf lamina). Further, to account for heterogeneous backgrounds and image artifacts, up to ten times as many background pixels are sampled from the interior of each leaf. The image tiles are augmented during training using a combination of random continuous rotations, horizontal and vertical flips, color augmentation, and Gaussian blur.

The vein growing CNN outputs $3\times3\times2$ predictions of the center pixel and its neighbors, in which the two prediction channels represent probabilities that a pixel belongs to the foreground (vein) or background (lamina). To measure error between predicted pixel probabilities and their ground truth classifications, Focal Loss ($\mathcal{L}_{\text{FL}}$), an extension of standard binary cross-entropy, is used as an objective function~\cite{lin2017focal}. In particular, Focal Loss seamlessly accounts for the class imbalance between positive and negative samples (i.e., there are many more background pixels than vein pixels) and allows the model to focus on more difficult examples where veins are obscured. The objective function is given by

\begin{equation}
    \mathcal{L}_{\text{FL}} = 
    \begin{cases}
        -\alpha \, (1-p)^\gamma \, \log(p) & \text{if $y=1$}
        \\
        -(1-\alpha) \, p^\gamma \, \log(1-p) & \text{otherwise},
    \end{cases} \label{eq:focalloss}
\end{equation}

\noindent where $p = \text{CNN}(x)$ are the pixel probabilities, $x \in \mathbb{R}^{128\times128\times3}$ is the input image tile, $y \in \{0, 1\}$ are the ground truth pixel classes, and $\alpha=0.25$ and $\gamma=2.0$ are the default hyperparameters of the Focal Loss function~\cite{lin2017focal}. {Vein growing CNNs are trained using both Focal Loss as well as binary cross-entropy and compared in Section~\ref{sec:results}.}

The model architecture and training strategy are nearly identical to the leaf tracing framework. Since the input tiles for vein segmentation are half the dimension of the inputs for leaf tracing, the first block of $3\times3$ convolutional layers and max pooling is removed from the architecture described in Section~\ref{sec:tracing}. Thus, the vein growing CNN transforms the spatial image dimensionality from $128\times128$ to $4\times4$, after which a final $4\times4$ convolution layer reduces the outputs to a vector of length 18, which is reshaped into $3\times3\times2$ for vein classification. Note that, unlike the leaf tracing CNN, the final convolution includes a Softmax activation function, which constrains the outputs to between 0 and 1 and motivates the probabilistic interpretation for the objective function. The model is trained with the Adam optimizer for 1,000 epochs with a batch size of 1024 and early stopping of 20. Note that a larger batch size is used here compared to the leaf tracer since the inputs are smaller and thus more can be included in each batch. Finally, six images (totaling $\sim$6.7M image tiles) are used for training and two (totaling $\sim$2.5M image tiles) for validation. 

Once the vein growing CNN is trained, it is used in a recursive framework in which the CNN decides whether new pixels should be added to the vein segmentation. The algorithm is initialized by randomly sampling 10,000 seed pixels inside the leaf body (using the segmentations from Section~\ref{sec:tracing}). For each seed pixel, image tiles are generated and fed to the model, which then classifies the seed pixel and its neighbors. Neighboring pixels that are classified as leaf veins are used as seeds in the next iteration. Once a seed pixel has been considered, it is removed from the sample set for future iterations. This process is then repeated, continuously adding pixels to the segmentation, until no new pixels are positively classified. Note that a pixel can receive multiple classifications as its neighbors become seeds during the iterations. To account for this, the final vein segmentation is determined by thresholding the \emph{average} probability of each pixel. The optimal probability threshold is chosen by minimizing the number of connected components in the segmentation mask across a range of threshold values. In other words, the optimal threshold is the one that maximizes vein connectivity in the segmentation mask. Like the leaf tracing framework, the vein growing algorithm does not require human supervision at inference time and completes accurate vein segmentations in {$\mathcal{O}(10)$} seconds on a single GPU of an NVIDIA DGX Station A100, which is orders of magnitude faster compared to human annotation.

\subsection{{Baseline comparison}} \label{sec:baseline}

\begin{figure}[h!]
    \centering
    \includegraphics[width=\textwidth]{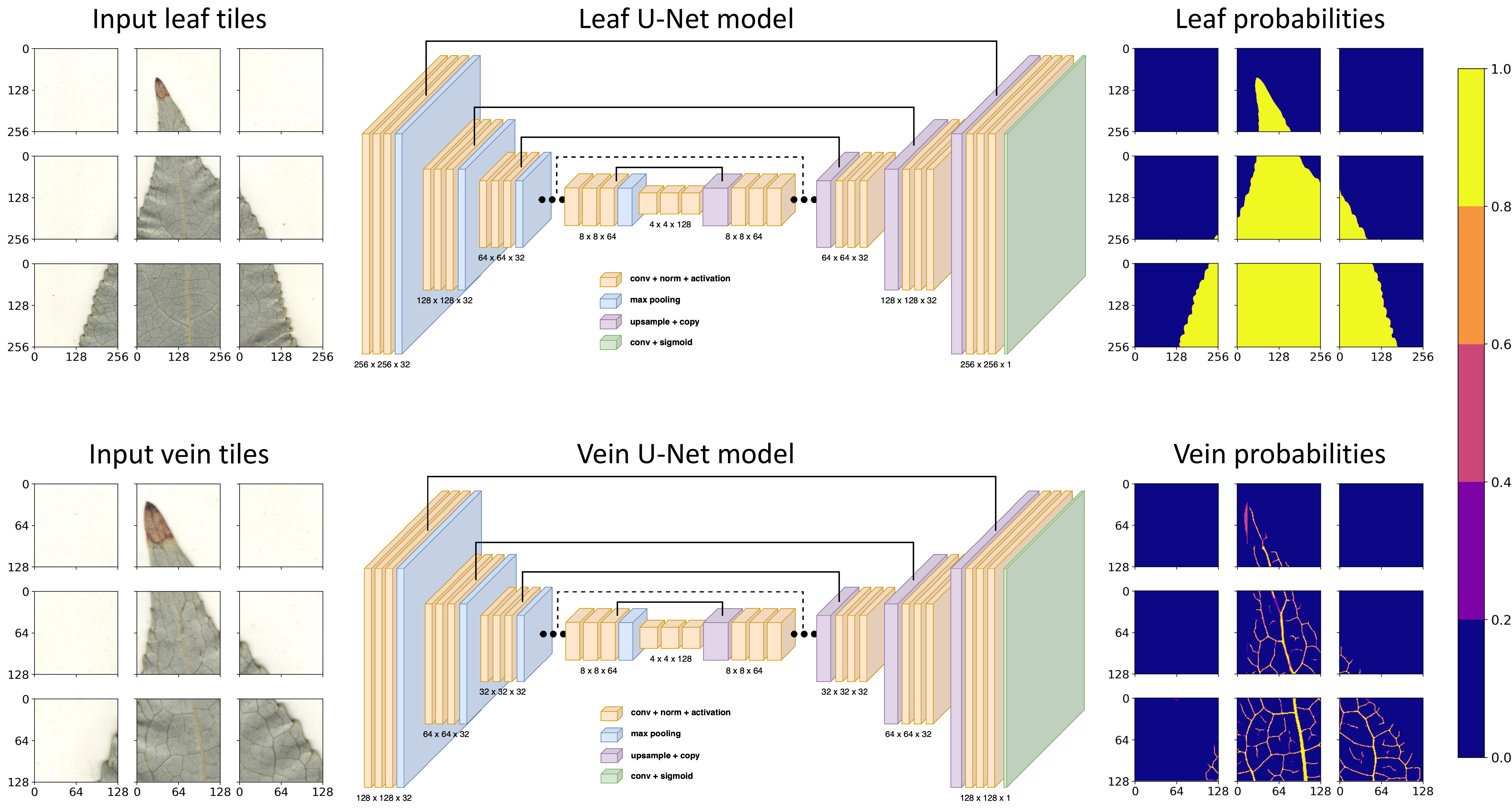}
    \caption{{\textbf{Baseline models.} U-Net is used as a baseline for comparison, in which large images are broken into smaller image tiles for prediction. Note that input tiles are sampled such that they overlap during prediction to account for edge effects. \underline{Top}: $256\times256\times3$ RGB input tiles are fed to a U-Net model which predicts a $256\times256$ matrix of leaf pixel probabilities. \underline{Bottom}: $128\times128\times3$ RGB input tiles are fed to a U-Net model which predicts a $128\times128$ matrix of vein pixel probabilities.}}
    \label{fig:unet}
\end{figure}

{A variant of the state-of-the-art image segmentation model, U-Net~\cite{ronneberger2015u}, is used as a baseline to compare against the leaf tracing and vein growing methods. Unlike the leaf tracing and vein growing CNNs, which consist of only an encoder network, U-Net uses an additional decoder network in an auto-encoder-like model structure. However, instead of the decoder network leveraging information from just the final layer of the encoder, U-Net includes skip connections from the encoder to the decoder at each spatial resolution so that the model may leverage multiple levels of information to accurately segment images. Following the same strategy for the leaf tracing and vein growing methods, U-Net is also trained on smaller image tiles which are then combined during prediction to reconstruct the image-level segmentation mask. See Figure~\ref{fig:unet} for a diagram of the model architectures used for leaf and vein segmentation.}

{Separate U-Net models are trained for leaf and vein segmentation. For leaf segmentation, following a similar procedure outlined in Section~\ref{sec:tracing}, U-Net inputs $256\times256\times3$ RGB image tiles that are $[0, 1]$ normalized and outputs corresponding $256\times256$ probability maps which predict pixels belonging to the leaf body. However, unlike the leaf tracing inputs, U-Net inputs are not constrained to be centered along the leaf boundary, and are instead sampled from anywhere in the leaf. For vein segmentation, following Section~\ref{sec:growing}, U-Net inputs $128\times128\times3$ RGB image tiles that are $[0, 1]$ normalized and outputs corresponding $128\times128$ probability maps which predict pixels belonging to the venation. Similarly, these input tiles are sampled from anywhere in the leaf. Each U-Net is trained using the same image augmentation techniques that are described in Sections~\ref{sec:tracing} and~\ref{sec:growing}.}

{The baseline models used in this work are variants of the original U-Net architecture described in~\cite{ronneberger2015u}, and are adapted from the CNN models described in Sections~\ref{sec:tracing} and~\ref{sec:growing}. In particular, each U-Net model uses an identical encoder to the leaf tracing and vein growing CNNs. However, rather than applying a final convolution layer to the encoder outputs, a decoder network is applied instead. The decoders use the same number of layers as the encoders, but with layer widths in reverse order (i.e., layer width increases with depth in the encoder, and decreases with depth in the decoder). Skip connections are drawn from the activation maps preceding each pooling layer in the encoder, and concatenated with the corresponding upsampled activation maps in the decoder. Upsampling is achieved using 2D transpose convolutions. The outputs of the last layer of the decoder are fed to a final convolution layer with a Sigmoid activation, which transforms the outputs into a 1-channel probability map used for pixel classification.} 

{To compare the models directly, each U-Net is trained using the same training and validation sets as the leaf tracing (i.e., 40 training images totaling $\sim$250K image tiles and 10 validation images totaling $\sim$50K image tiles) and vein growing (i.e., six training images totaling $\sim$6.7M image tiles and two validation images totaling $\sim$2.5M image tiles) models. The models are trained for 1,000 epochs with early stopping set to 20 iterations. However, due to memory constraints (i.e., U-Net uses approximately twice the memory compared to the leaf tracing and vein growing CNNs), half of the batch size is used for the respective segmentation tasks (i.e., 128 for leaf segmentation and 512 for vein segmentation). Following Section~\ref{sec:growing}, U-Net is trained using binary cross-entropy for leaf segmentation and binary cross-entropy and Focal Loss for vein segmentation, which are compared in Section~\ref{sec:results}.}

{For whole-image segmentation of the leaf body and vein architecture, each image is broken into uniform grid of small overlapping image tiles. For leaf tracing, $256\times256$ tiles are sampled with a step size that is half of the window size (i.e., 128). Similarly for vein segmentation, $128\times128$ tiles are sampled with a step size of 64 pixels. In this way, each pixel is considered from multiple positions in the image, thereby reducing poor predictability near the edges where the network has less context. After all of the image tiles have been predicted by U-Net, the average probability is computed for each pixel and then thresholded to produce the final segmentation mask. For leaf segmentation, the probability threshold is set to 0.5. For vein segmentation, the optimal probability threshold is chosen to minimize the number of connected components in the vein structure in order to maximize vein connectivity. Like the leaf tracing and vein growing methods, the U-Net prediction algorithm requires no human interaction, and completes segmentations in $\mathcal{O}(1)$ seconds for leaf segmentation and $\mathcal{O}(10)$ seconds for vein segmentation on a single GPU of an NVIDIA DGX Station A100. Note that for vein segmentation, though both methods segment images in the same order of magnitude of seconds, U-Net is consistently 2-3 times faster owing to the smaller number of image tiles the model needs to predict (e.g., 20 seconds rather than 60 seconds per image).}

\subsection{Feature extraction} \label{sec:featureextraction}

Given the binary segmentation maps from Sections~\ref{sec:tracing} and~\ref{sec:growing}, traditional open-source image-processing tools are used to extract biologically meaningful traits from the leaf body, vein architecture, and petiole. This is possible since the segmentation methods effectively remove background artifacts and highlight the salient information in leaf scans. In this work, Fiji~\cite{schindelin2012fiji} is used to extract leaf-level traits, RhizoVision Explorer (RVE)~\cite{seethepalli2020rhizovision, seethepalli2021rhizovision} is used for vein traits (e.g., length and thickness), and a custom implementation is used for petiole traits (length and width). RVE is chosen for vein traits in particular since it is designed to analyze root systems, which are composed of vessel-like structures with tips, branch points, redundant connections, etc., and makes it applicable to studying vein architectures, which share many of the same characteristics. Further, since the scan resolution is known (i.e., 300 DPI), features extracted from Fiji and RVE are easily converted from pixel-coordinates to standard units (e.g., cm). 

Fiji is applied to the leaf segmentations from Section~\ref{sec:tracing} to extract 23 image-based traits related to whole-leaf morphology. Some morphological descriptors include area (cm$^2$), perimeter (cm), circularity (unitless), and solidity (unitless), etc. Color features are also derived by relating the segmentations back to the original scanned images, including average red, green, blue, hue, saturation, and brightness values corresponding to leaf pixels. Feature extraction in Fiji is scripted and applied in ``batch mode'' to the full set of leaf segmentations. A detailed description of each of leaf-level trait is provided in Supplementary Table~\ref{tab:leaftraits}.

RVE is used to extract 27 features from the vein segmentations. Note that only vein pixels inside the leaf segmentation were used for vein architecture traits (i.e., the petiole is not considered here). The software parameters are set to 300 DPI and ``whole mode'' for image-level traits. Vein diameter ranges are used classify veins into three ranges: (i) less than 0.25mm, (ii) between 0.25mm and 0.80mm, and (iii) above 0.80 mm, in an attempt to correspond to third, second, and first order veins, respectively. Extracted traits include those supplied by default (e.g., average vein diameter (mm), length (mm), and area (mm$^2$)), with some traits being repeated across the three vein diameter ranges. Following~\cite{sack2013venation}, additional venation traits are also derived that measure proportions between vein length/area to leaf morphology. See Supplementary Table~\ref{tab:veintraits} for the full list of vein traits and their descriptions.

Petiole segmentations are derived by considering the largest connected component of vein pixels outside of the leaf segmentation. Then, to compute petiole length and width, Fiji is used to compute the best-fit rotated rectangle around the petiole mask. The height of the bounding rectangle is sufficient to estimate petiole length. However, rectangle width is not used to estimate petiole width since (i) petiole width changes along the length of the petiole (i.e., it tends to be wider near the ends and thinner near the midpoint), and (ii) the caliper measurements for petiole width were taken near the center of the petiole. Thus, petiole width is estimated by computing the average diameter over the center 20\% of the segmentation. Finally, Fiji is used to estimate similar traits for the petiole compared to the leaf body (e.g., area and perimeter), and RVE was used to estimate petiole volume. See Supplementary Table~\ref{tab:petioletraits} for the full list of 18 petiole traits and their descriptions. 

The feature extraction process yields 68 traits related to leaf, vein, and petiole morphology that can be used for genomic analysis. To validate image-based features with real-world measurements, petiole length and width are compared against caliper measurements that were recorded manually during image capture. To consider the results from a biological perspective, (i) broad-sense clonal heritability is computed for each recorded trait, and (ii) a genome-wide association study (GWAS) is performed for the vein density trait (i.e., the ratio of vein area to leaf area). Vein density is chosen since it utilizes both the leaf and vein segmentations, and since the ratio between lamina and venation must balance sunlight intake and carbon fixation with the transport of sugars and other nutrients to sink organs, all of which are essential processes for biomass production.

\subsection{Validation} \label{sec:validation}

To validate the leaf and vein segmentations, following~\cite{rutter_tracing} and~\cite{lagergren_growing}, the Jaccard index (intersection over union) is used to measure segmentation accuracy for images in the validation sets (i.e., ten for leaf segmentation and two for vein segmentation). This metric measures similarity between two semantic segmentations by computing the ratio between the set intersection (all true positive pixels) and the set union (all true positive, false positive, and false negative pixels), where scores near one indicate high accuracy and near zero indicate low accuracy. {These values are compared between the leaf tracing CNN, vein growing CNN, and baseline U-Net models.}

{To account for cases in which set similarity metrics like the Jaccard index may be misleading, an additional perspective is taken to measure the biological accuracy of the vein segmentations. In particular, the leaf vasculature is made up of a single connected network of veins. In other words, a perfect segmentation mask of the vasculature should, in theory, contain just one connected component. In practice, this is exceedingly difficult to obtain since veins are only partially observable from the surface of the leaf lamina (i.e., veins appear and disappear throughout the leaf surface), meaning that even the ground truth segmentations include multiple connected components. Despite this observation, an accurate vein segmentation must not only overlap with the ground truth mask (i.e., maximize the Jaccard index) but must also be biologically realistic (i.e., minimize the number of connected components). Thus, in addition to computing the Jaccard index for vein segmentations, the number of connected components are also computed and compared across the vein growing CNN and U-Net baseline models.} 

Note that since validation error was monitored during training, conclusions drawn from segmentation accuracy for validation images may be affected by data leakage (i.e., create an over-optimistic interpretation of the model). To account for this, the predicted digital measurements across the population are further validated using real-world physical measurements. To this end, calipers were used to measure petiole length and width during data collection. These values are compared against the corresponding features extracted from the vein segmentations described in Sections~\ref{sec:growing} and~\ref{sec:featureextraction}. To measure the agreement between digital and manual values quantitatively, the coefficient of determination ($R^2$) is computed for each trait.

\subsection{Genomic analysis} \label{sec:genomics}

To pre-process the vein density trait for GWAS, outliers are removed using median absolute deviation (MAD), where any measurement with $\text{MAD} > 6$ is removed. To account for geospatial variation across the plantation, thin plate spline (TPS) correction is applied using the \textit{fields} software package in R~\cite{fieldsR}, in which the row and position of each tree are used as coordinates for the TPS models. To extract the genetic component of each sample, best linear unbiased predictors (BLUPs) are computed for the TPS-corrected values using the \textit{lme4} software package in R~\cite{bateslme4}, which fits genotypes as random effects for each trait. In addition, to assess the repeatability of each measurement and the genetic control of the vein density trait, broad-sense heritability ($H^2$) was estimated using the TPS-corrected values of the clonal replicates (131 replicated samples) from the two blocks considered in this work. Heritability is computed by

\begin{equation}
    H^2 = \frac{\sigma^{2}_{G}}{\sigma^{2}_{G} + \sigma^{2}_{E}}, \label{eq:genomicvariance}
\end{equation}

\noindent where $\sigma^{2}_{G}$ is the genotypic variance due to clonal differences and $\sigma^{2}_{E}$ represents environmental variance.

For genomic analysis, a total of 1,492 \textit{P. trichocarpa} accessions were previously sequenced using the Illumina genetic analyzer with paired-end sequencing technology at the Department of Energy Joint Genome Institute~\cite{nordberg2014genome}. The sequences are aligned to the v4 reference genome using the Burrows-Wheeler Alignment tool, BWA-MEM~\cite{li2013aligning}, and variant calling is performed using the GATK (version 4.0) Haplotype caller~\cite{van2013fastq}. Starting with more than 22 million single nucleotide polymorphisms (SNPs) obtained by the GATK Variant Quality Score Recalibration (VQSR) method at tranche 99, 847,066 SNPs across 1,419 genotypes were retained for population-scale genomic analysis after applying the following filters. 73 individuals were removed due to having excessive genomic relatedness or having greater than 10\% missing SNP data. SNPs were removed if they had greater than 15\% missing genotypes, or minor allele frequency less than 0.05, or Hardy Weinberg Equilibrium chi-square test P value $< 10^{-50}$. SNPs were further pruned using a linkage disequilibrium (LD) coefficient of determination threshold of $R^2 \geq 0.7$.

The data pre-processing steps above yield 847,066 SNPs for 1,419 unrelated genotypes that are used for GWAS analysis of the vein density trait. Association between the SNPs and the phenotypic vector was tested using a multilocus GWAS method, BLINK, from the GAPIT3 software package in R~\cite{huang2019blink}, that uses two fixed effect models (FEM) iteratively. The first FEM tests for the association of all genetic markers independently to generate a set of pseudo Quantitative Trait Nucleotides (QTNs) that are then used in the second FEM to optimize the selection of pseudo QTNs. Only those QTNs that are significant and not in LD are used as covariates in the association test. The first FEM is given by 

\begin{equation}
    y_i = S_{i1}b_1 + S_{i2}b_2 + \cdots + S_{ik}b_k + S_{ij}d_j + e_i, \label{eq:fem1}
\end{equation}

\noindent where $y_i$ is the phenotypic value of the $i^{\text{th}}$ individual, $S_{i1}, \dots, S_{ik}$ are the genotypes of the $k$ QTNs, $b_1, \dots, b_k$ are the corresponding effects of the QTNs, $S_{ij}$ is the genotype of the $i^{\text{th}}$ individual and $j^{\text{th}}$ SNP, $d_j$ is the $j^{\text{th}}$ SNP effect, and $e_i$ is the residual. The second FEM is used to optimize the QTNs for use as covariates in the first FEM, and is given by

\begin{equation}
    y_i = S_{i1}b_1 + S_{i2}b_2 + ... + S_{i}b_k + e_i, \label{eq:fem2}
\end{equation}

\noindent with a similar interpretation to Equation~\ref{eq:fem1}. Note that Equation~\ref{eq:fem2} is essentially a reduced version of Equation~\ref{eq:fem1}, in which the SNP term that tests for the association with the phenotypic vector is removed. The model optimization is performed with Bayesian Information Criterion (BIC).


\section{Results} \label{sec:results}

\subsection{Segmentation results}

\begin{figure}[h!]
    \centering
    \includegraphics[width=\textwidth]{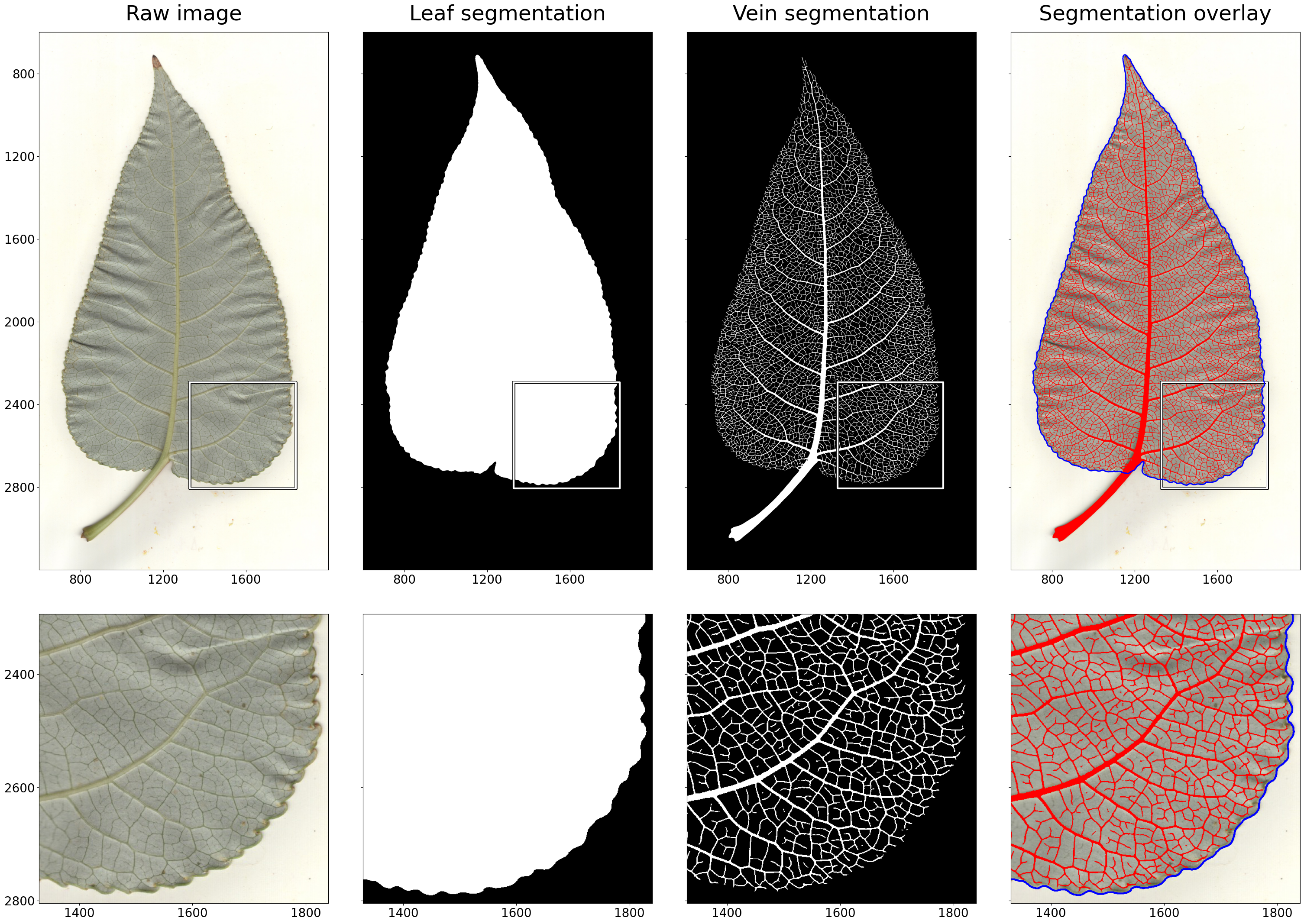}
    \caption{\textbf{Leaf and vein segmentations.} Results of the leaf and vein segmentation methods on an example leaf outside the training set. The top row shows the full leaf and the bottom row gives a zoomed in view. \underline{Left}: example leaf scan chosen from outside the training and validation sets. \underline{Center left}: predicted segmentation of the leaf body {using the leaf tracing CNN}, where pixels inside the traced contour are shown in white and outside the contour in black. \underline{Center right}: predicted segmentation of the visible vein architecture {using the vein growing CNN with Focal Loss}, where vein pixels are shown in white and background pixels in black. \underline{Right}: example leaf scan with the predicted leaf boundary and vein architecture overlaid in blue and red, respectively. Note that for visualization these images are zoomed in to remove redundant white space from the scanner background.}
    \label{fig:segmentation}
\end{figure}

The few-shot learning {and baseline} methods are applied to the total set of images, in which the 2,906 top and bottom scans are used for leaf segmentation, and the 1,453 bottom scans are used for vein architecture. Examples of the resulting model outputs are given in Figure~\ref{fig:segmentation}. Note that the leaf in Figure~\ref{fig:segmentation} was not used for model training or validation. Additional segmentation results are visualized in Supplementary Figure~\ref{fig:overlays}, which illustrates leaf heterogeneity by varying leaf size and vein density. All of the image data, ground truth annotations, and predicted leaf/vein segmentations {are made publicly available at~\cite{fsl_code, fsl_data}}.

The Jaccard index is used to measure segmentation accuracy for images in the validation sets (i.e., ten for leaf segmentation and two for vein segmentation), where scores near one indicate high accuracy and near zero indicate low accuracy. {For leaf segmentation, the leaf tracing CNN achieves a mean ($\pm$ std) Jaccard score of 0.9946 ($\pm$0.0016) while the baseline U-Net model achieves a score of 0.9969 ($\pm$0.0009)}, indicating a high degree of overlap between the predicted and ground truth segmentations {and virtually no difference in accuracy between the two methods}. For vein segmentation, the two validation images {are used to compare Jaccared scores between the vein growing CNN and U-Net models  and binary cross-entropy (BCE) and Focal Loss (FL) objective functions, and are given in Table~\ref{tab:jaccard_scores}. Note that to compare each method equally, the Jaccard scores are computed using using a probability threshold of 0.5 for each case. This is in contrast to the other results and visualizations, which use thresholds that minimize the number of connected components in each segmentation mask.}

\begin{table}[h!]
    \centering
    \caption{{\textbf{Vein segmentation accuracy.} Jaccard index is used to quantify vein segmentation accuracy for the two images held out for validation (see the top of each table for the leaf ID). The vein growing CNN (Grower) is compared to the baseline model (U-Net) for both binary cross-entropy (BCE) and Focal Loss (FL). The model/loss combination resulting in the largest Jaccard score is shown in bold.}}
    {\begin{tabular}{c|cc}
        \multicolumn{3}{c}{\url{C_1_14_18_bot.png}} \\ 
               & BCE    & FL              \\ \hline
        U-Net  & 0.6312 & 0.5947          \\
        Grower & 0.5990 & \textbf{0.6586} \\
    \end{tabular}}
    \hspace{2em}
    {\begin{tabular}{c|cc}
        \multicolumn{3}{c}{\url{C_1_8_1_bot.png}} \\ 
               & BCE    & FL              \\ \hline
        U-Net  & 0.5965 & 0.5633          \\
        Grower & 0.6047 & \textbf{0.6381} \\
    \end{tabular}}
    \label{tab:jaccard_scores}
\end{table}

The reduced scores for vein segmentation compared to leaf segmentation is due mainly to (i) human errors in the ground truth segmentation and (ii) the complexity of the vein architecture. For example, the model identifies veins that were missed during manual annotation, and thus are considered false positives in the Jaccard score. Further, due to thin veins, predicted veins that are off by just one pixel can result in large changes in Jaccard score. For a visualization of these phenomena, see Supplementary Figure~\ref{fig:accuracy}, which illustrates these effects for the validation image with the lowest Jaccard score. Despite the lower Jaccard metric, the vein growing framework achieves recall/sensitivity values (i.e., the probability of detecting a vein pixel) of 0.9219 and 0.8673, respectively, which indicates that the method has a high detection rate, and thus almost completely captures the structure of the visible vein architecture. 

\begin{figure}[h!]
    \centering
    \includegraphics[width=0.85\textwidth]{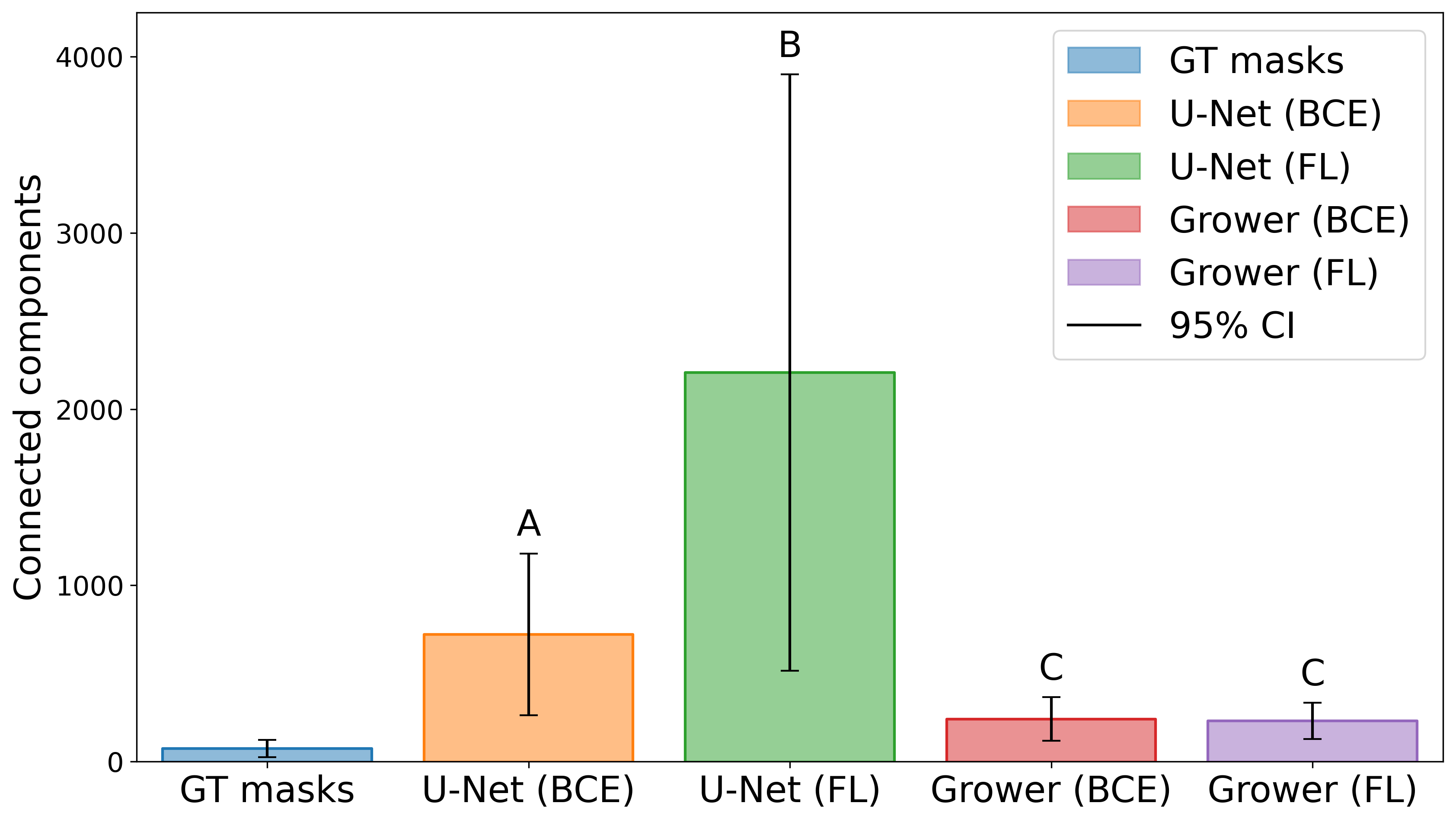}
    \caption{{\textbf{Vein object counts.} Biological accuracy of the vein segmentations is measured by counting the number of connected components in the eight ground truth (GT) and sets of 1,453 predicted masks. The vein growing CNN (Grower) is compared against the baseline model (U-Net) trained using binary cross-entropy (BCE) and Focal Loss (FL). The vertical axis represents the number of connected components in the vein segmentations. The height of each bar corresponds to the mean object count, with black lines indicating 95\% confidence intervals (CI). The results of a Tukey's honestly significant difference (HSD) test groups the means of the four models into three statistically significant groups: A, B, and C. Note that ground truth masks are not considered in this analysis due to the orders-of-magnitude difference in number of samples.}}
    \label{fig:vein_objects}
\end{figure}

{In addition to Jaccard index, the number of connected components in each vein segmentation is computed as a proxy for biological accuracy. In other words, since the vein architecture of a leaf is a single fully-connected network, a biologically accurate vein segmentation should minimize the number of connected components. To this end, bar plots are constructed for the vein segmentations arising from the vein growing and baseline CNN models, and are visualized in Figure~\ref{fig:vein_objects}. Note that vein objects are counted using custom probability thresholds that minimize the number of connected components in each image for each model as described in Sections~\ref{sec:growing} and~\ref{sec:baseline}.}

{Figure~\ref{fig:vein_objects} demonstrates that the vein growing CNN, while producing more fragmented segmentations compared to the ground-truth masks, significantly reduces the number of vein objects compared to U-Net, regardless of objective function. Further, while the objective function does not significantly change the number of connected components for the vein growing CNN, there is a drastic difference between U-Net models. To quantify these observations, the mean number of connected components of each model's outputs is compared using Tukey's honestly significant difference (HSD) test with $\alpha=0.05$. This test reveals that the four models are reduced to three statistically significant groups, A: U-Net with BCE, B: U-Net with FL, and C: vein growing CNN with BCE and FL. These findings show that the three groups differ statistically significantly in value, with the vein growing CNN producing the smallest number of vein objects. Thus, since the vein growing CNN with FL results in the highest segmentation accuracy (see Table~\ref{tab:jaccard_scores}) and approximately the same biological accuracy to BCE (see Figure~\ref{fig:vein_objects}), it is chosen as the optimal model for vein segmentation in this work. In particular, this model is used for feature extraction, validation with physical measurements, and downstream genomic analysis.}

\begin{figure}[h!]
    \centering
    \includegraphics[width=0.85\textwidth]{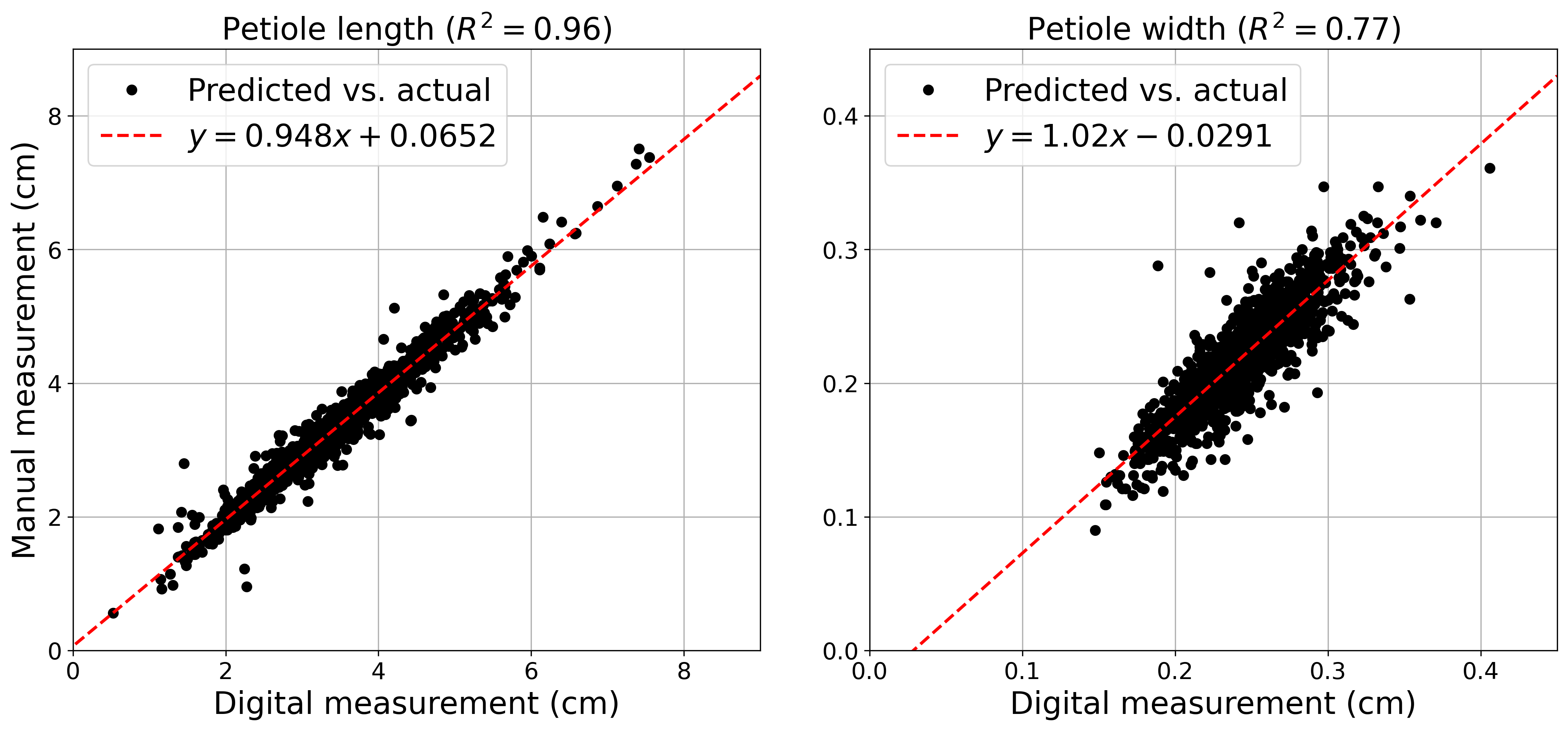}
    \caption{\textbf{Petiole measurement validation.} Each subplot compares predicted ($x$-axis) with actual ($y$-axis) morphological measurements of the petiole. Manual measurements were obtained with calipers during data collection while digital measurements were derived from the leaf and vein segmentations. \underline{Left}: petiole length comparison with data shown in black and the best-fit line shown in red. \underline{Right}: petiole width comparison with data shown in black and the best-fit line shown in red.}
    \label{fig:validation}
\end{figure}

The predicted digital measurements across the population are further validated using real-world caliper measurements, which are visualized in Figure~\ref{fig:validation}. In particular, the data are compared against a linear model, which results in $R^2 = 0.96$ for petiole length and $R^2 = 0.77$ for petiole width. This discrepancy between $R^2$ values is due to several factors. First, manual measurement of petiole length is made from end to end, resulting in larger, more consistent measurements. However, for petiole width, the caliper was placed at the approximate center of the petiole, resulting in greater variation due to subjective positioning of the caliper. Further, since petiole width is typically smaller, measurement errors affect the $R^2$ value more compared to petiole length. Despite these effects, the digital traits strongly agree with the manual measurements, which further helps validate the accuracy of the segmentations and extracted features.

\subsection{Genomic analysis results} 

\begin{figure}[h!]
    \centering
    \includegraphics[width=\textwidth]{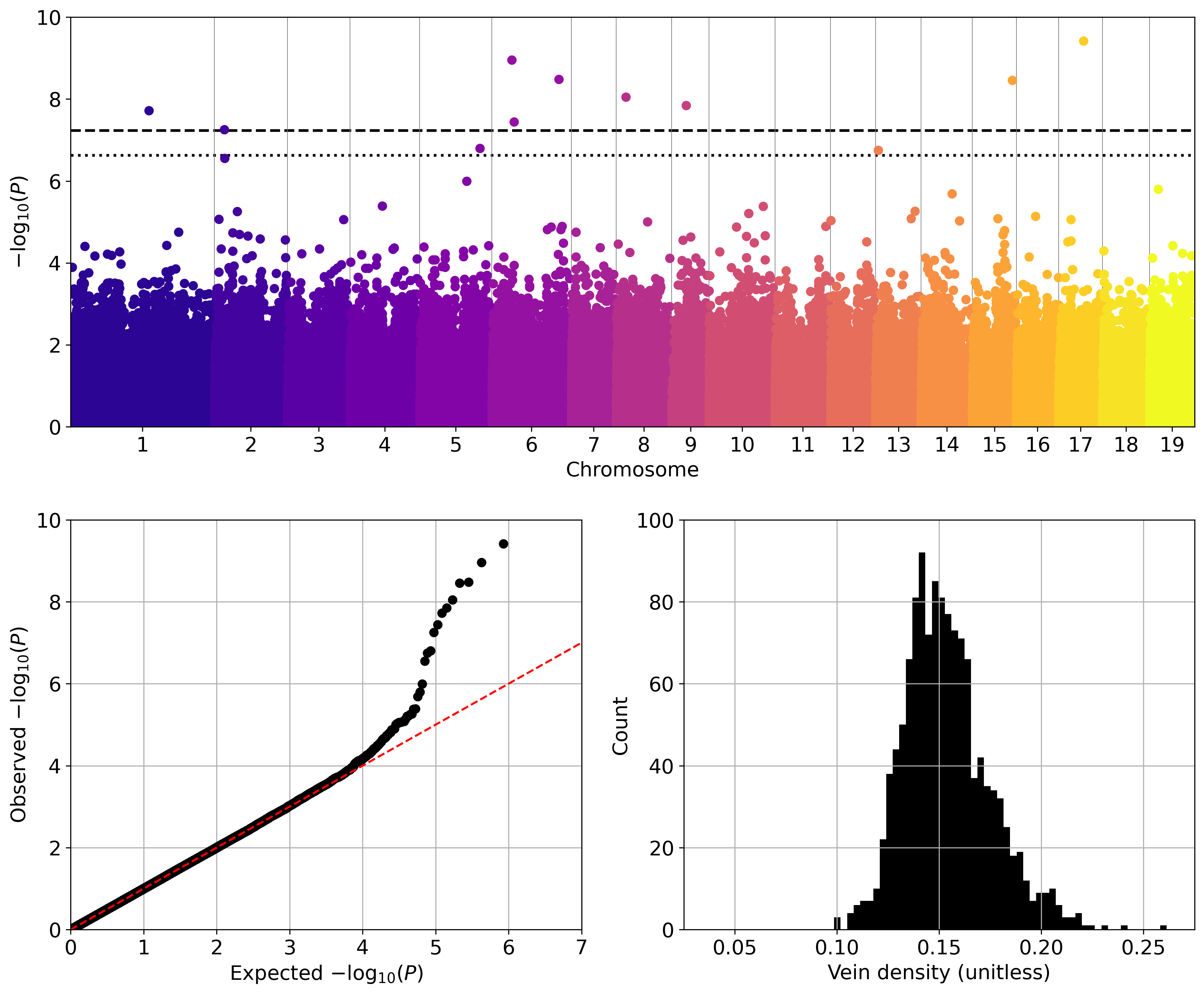}
    \caption{\textbf{Population-scale genomic analysis.} \underline{Top}: a Manhattan plot of the GWAS results for leaf vein density using the multilocus BLINK method for 847,066 SNPs across 1,419 \textit{P. trichocarpa} genotypes. The horizontal axis corresponds to genomic positions by chromosome and the vertical axis shows the negative log-base-10 P value for each SNP. The dashed horizontal line represents the FDR threshold, $P<0.05$, and dotted line represents the FDR threshold, $P<0.2$. \underline{Bottom left}: the quantile-quantile (QQ) plot corresponding to the P values shown in the Manhattan plot with the expected values shown by the red dashed line. \underline{Bottom right}: the distribution of TPS-corrected vein density BLUPs used for GWAS.}
    \label{fig:gwas}
\end{figure}

To consider the segmentation and feature extraction methods from a biological perspective, a GWAS analysis is conducted at population-scale to associate vein density (i.e., the ratio of vein area to leaf area) to the \textit{P. trichocarpa} genome. The broad-sense clonal heritability of the vein density trait is moderately high ($H^2 = 0.65$), which suggests that the trait is under genetic control. The multilocus BLINK method is used to perform GWAS on the vein density trait with 847,066 SNPs in the genome. This analysis identified 12 significant SNPs using a false discovery rate (FDR) P value, $P<0.05$, and 15 unique SNPs using a less-strict FDR P value, $P<0.2$. A Manhattan plot, quantile-quantile (QQ) plot, and the distribution of the TPS-corrected vein density BLUPs are shown in Figure~\ref{fig:gwas}. A total of 30 unique genes that potentially control for the variation in vein density trait are identified for these SNPs based on the nearest flanking genes in both directions of the GWAS hits in the genome. To gain more insight into the function of these genes, the \textit{Arabidopsis thaliana} orthologs are identified based on the protein sequence similarity. The GWAS results are summarized in Table~\ref{tab:gwas}. See Section~\ref{sec:discussion} for discussion about these genes and their associated physiological plant processes.


\begin{table}[h!]
    \centering
    \caption{\textbf{Identified genes.} Top gene models detected by the BLINK GWAS method based on FDR  P value $P<0.2$ for vein density in \textit{Populus trichocarpa}. Each row corresponds to different genes and includes the gene ID, chromosome number, SNP position, distance in the genome (positive for upstream and negative for downstream from the SNP position), minor allele frequency (MAF), P value, \textit{A. thaliana} ortholog, and ortholog annotation.} 
    \label{tab:gwas} 
    \begin{tabular}{ccccccp{2.3cm}}
        \hline
        \textbf{Gene ID} & \textbf{Chr. (pos.)} & \textbf{Dist.} & \textbf{MAF} & \textbf{P value} & \textbf{Ortholog} & \textbf{Annotation} \\
        \hline
        Potri.017G077200  & 17 (8,592,229)  & -5,329 & 0.1246 & 3.24e-4 & AT3G04680 & CLP-SIMILAR PROTEIN 3 \\
        Potri.017G077300  & 17 (8,592,229)  &  6,036 & 0.1246 & 3.24e-4 & AT3G01300 & PBS1-LIKE 35 \\
        Potri.006G090501  &  6 (6,910,580)  & -2,825 & 0.1938 & 4.69e-4 & -         & - \\
        Potri.006G090600  &  6 (6,910,580)  &  1,617 & 0.1938 & 4.69e-4 & AT3G53880 & ALDO-KETO REDUCTASE FAMILY 4 MEMBER \\
        Potri.006G227300  &  6 (23,159,363) & -3,829 & 0.2158 & 7.47e-4 & AT1G18600 & RHOMBOID-LIKE PROTEIN 12 \\
        Potri.006G227400  &  6 (23,159,363) &  2,083 & 0.2158 & 7.47e-4 & AT3G13784 & CELL WALL INVERTASE 5 \\
        \hline
    \end{tabular}
\end{table}


\section{Discussion} \label{sec:discussion}

In this work, few-shot learning was used for both leaf and vein segmentation. In particular, each method divided a small number of large images into a large number of small image tiles and used the predictions of previous iterations to expand partial segmentations until stopping criteria was reached. In particular, iterative data set refinement was paired with heavy image augmentation so that the CNN models learned an invariance to potential image artifacts not included in the small number of annotated images. In this way, the complex task of whole-image segmentation was broken down into smaller easier decision rules, thereby maximizing predictive accuracy while minimizing the amount of labeled images needed for model training. This strategy was chosen primarily due to (i) the lack of labeled training images (e.g., only 50 leaf segmentations and eight vein segmentations) and (ii) because each leaf scan is large ($3510 \times 2550$ pixels) and does not easily fit into a standard deep learning model (e.g., a CNN). To address (i), one could simply annotate more data manually, however, to segment the visible venation in the images considered here (Section~\ref{sec:data}) would require up to 12,000 person-hours (i.e., 6 years, assuming a 40-hour work week). Note that the approaches discussed in this work could be used to automatically label training data for larger deep learning workflows (e.g., U-Net~\cite{ronneberger2015u} or Mask-RCNN~\cite{he2017mask}). For (ii), large images are typically greyscaled, reshaped, or downsampled to fit within system requirements and hardware limitations~\cite{xu2021automated}. However, this strategy can alter or destroy fine-grained details that may be crucial for accurate prediction. Thus, the methods demonstrated here utilized raw RGB images at full resolution, but rather than using entire images as inputs, smaller tiles were sampled from within images to iteratively segment leaf boundaries and visible venation. For example, see Supplementary Figure~\ref{fig:overlays}, which illustrates how these tile-based approaches are generally insensitive to changes in object size and characteristics. 

Leaf segmentation was formulated as a tracing task, in which a CNN inputs image tiles centered along the boundary of a leaf, and outputs trace predictions that are used to sample new tiles in the next iteration. This methodology performed well in this application since there is only one leaf per image and each leaf is fully contained within the image. Note that images with multiple objects have been previously considered~\cite{rutter_tracing, rutter_combo}, but additional modifications to the algorithm (e.g., recurrence) may be needed to account for images with cluttered objects and overlapping boundaries. The resulting leaf segmentations were highly accurate and captured the morphology and serration of each leaf. Since each leaf was scanned against a white background, automatic thresholding could be used to obtain a rough segmentation of the leaf. However, this approach captures background artifacts, includes the petiole, and falsely detects shaded regions near the leaf boundaries and petiole. These artifacts could be addressed for individual samples by post-processing the binary segmentation maps (e.g., binary erosion and dilation), but defining such rules that generalize to all cases across the population is increasingly difficult. In contrast, a strength of the tracing approach is that the tracing CNN can \emph{learn} an invariance to image artifacts in a data-driven way without human supervision, and still leverage the auto-threshold segmentations for trace initialization, which allows the method to be fully automated. Finally, the tracing methodology produced accurate leaf segmentations approximately three orders of magnitude faster than human annotation ({$\mathcal{O}(1)$} second compared to 15-30 minutes per image), allowing the method to scale up to population-level data sets, especially for computing systems that support parallelization (i.e., tracing more than one image at a time).

Due to the complexity of leaf venation, vein segmentation was formulated as a region growing task where a CNN predicts whether to include pixels in a segmentation by inputting image tiles centered at those pixels. Unlike the tracing framework (which segments objects by tracing boundaries in 1D), the region growing approach grows the segmentation directly by continuously adding pixels to the region of interest in 2D (see~\cite{januszewski_floodfilling} for 3D). A strength of this approach is that each pixel is considered individually. However, unlike previous methods which conduct an exhaustive prediction over all pixels in an image (which would equate to 8,950,500 pixels per image in this work)~\cite{ciresan2012deep}, pixels were only considered if they exceed a probability threshold, which allowed the model to focus only on pixels of interest. This distinction dramatically increased segmentation speed ({$\mathcal{O}(10)$} seconds compared to 4-8 hours per image with manual segmentation). In addition, the region growing framework can use a random sample of pixels from anywhere in the image to initialize the iteration. Note that in this work, seed pixels were drawn using the leaf body segmentations from the tracer to reduce the number of redundant white background pixels. Finally, and perhaps most importantly, the model produced accurate segmentations at population scale using just six images for training and two images for validation. This is in contrast to previous approaches that used CNNs for vein segmentation and required more than 700 ground truth vein segmentations~\cite{xu2021automated}. In particular, this result highlights the importance of iterative data set refinement, in which images were specifically added to the training set in order to build an invariance to observed artifacts in the population (e.g., leaf folds that were falsely classified as veins), which has also been noted in similar applications for root segmentation~\cite{smith2022rootpainter}.

Leaf and vein segmentation were specified as independent tasks, and used separate computational strategies to achieve each goal. Since both approaches were designed for segmentation, a natural question arises concerning whether two distinct approaches are necessary, or whether one would suffice for both tasks. In principle, the region growing CNN could be used for leaf segmentation, however, this would be computationally inefficient due to the large number of leaf pixels in the high-resolution scans. Compared to the tracing CNN (which focuses solely on boundary pixels), the region growing CNN would waste a large amount of computational resources on redundant ``interior'' pixels, which vastly outnumber boundary pixels. In the reverse case, the tracing CNN could theoretically be applied to vein segmentation. However, since the vein architecture is not homotopic to a circle, the tracing CNN would need to be re-initialized thousands of times inside the leaf to account for all of the ``holes'' in the vein architecture. Further, overlapping tracer boundaries near one-pixel-thick veins would create additional challenges in post-processing the thousands of traced contours. {Thus, iterative methods of this kind may only be effective for certain types of segmentation tasks. In particular, images containing multiple objects that overlap and obscure each other may be less suitable for tracing, while high-resolution images with large objects may be computationally infeasible for region growing. Therefore, it is important to account for such nuances when selecting a method for a particular task.} For example, in this work, each method was particularly well-suited for the segmentation task it was assigned. 

{The leaf tracing and vein growing CNNs were compared with a variant of the state-of-the-art image segmentation model, U-Net. Almost every aspect of the models and training strategies were kept equal, e.g., identical encoder networks, same data sets, input sizes, augmentation techniques, loss functions, training parameters, thresholding techniques, etc., with only a small number of changes to account for fundamental differences between the models, e.g., output size and batch size. The U-Net models provided multiple strengths. In particular, whereas tracing and region growing approaches are applied separately to leaf and vein segmentation, U-Net is agnostic to these objectives and only differs in the size of the input/output resolution. Further, U-Net reported approximately equal segmentation speed and accuracy compared to the leaf tracing CNN. For vein segmentation, U-Net segmented images in the same order of seconds, but consistently 2-3 times faster than the vein growing CNN. However, U-Net requires more than two times the memory per sample compared to the methods presented here (296.30 MB compared to 139.91 MB for leaf segmentation and 76.93 MB compared to 36.33 MB for vein segmentation), which may limit its use on smaller compute hardware (e.g., edge-computing devices for field deployment). Additionally, U-Net reported lower Jaccard scores and dramatically more fragmented segmentations compared to the vein growing CNN, which were quantified using Tukey's HSD test. Thus, the vein growing CNN is a more biologically accurate and memory-efficient model for vein segmentation, with important implications for downstream scientific analyses where vein connectivity is important (e.g., leaf hydraulic modeling, gas exchange, length/branching estimation, biomechanics, etc.). Note that even the ground truth segmentations included more than one connected component, since veins appear and disappear throughout the leaf lamina, which necessitates further development of methods to infer these missing connections in future work.}

A utility of the segmentation methods discussed here is that they remove background artifacts and highlight salient information in images (e.g., the leaf body or vein architecture). Using traditional computer vision applications (e.g., Fiji and RVE) to extract digital traits from such segmentations becomes trivial compared to using the raw image data. These advances not only reduce human effort, but also expand the number and variety of traits one can extract by including traits that can only be estimated digitally (e.g., vein density, leaf solidity, etc.). Further, custom algorithms can be developed that extract cryptic phenotypes related to leaf morphology and topological information from the vein networks which may yield new biological insights into the role leaves play in plant physiology -- this analysis is left for future work. These methods were also used to estimate measurable traits like petiole width and length, which were used as a source of validation in this work. This study demonstrated that manually estimated petiole length and width strongly correlated with their corresponding digital measurements, suggesting that the segmentation quality and feature extraction methodology accurately predicts biologically relevant features from the raw images. A strategy to consider these digital traits from a biological perspective was to estimate their broad-sense heritability (i.e., the amount of variation in the trait that is controlled by genetics), denoted by $H^2$. In particular, the $H^2$ value for vein density was 0.65, suggesting that the trait is under significant genetic control.

As a proof of concept for downstream application of the segmentation methods, a GWAS analysis was performed for the vein density trait at population scale. The top GWAS hit was Potri.017G077200, which is highly expressed in apical bud, dormant bud, and stem~\cite{sreedasyam2022jgi}. This gene is also expressed in immature/young leaves, suggesting that it plays a role in such tissues~\cite{sreedasyam2022jgi}. Comparing to \textit{A. thaliana}, \textit{CLP-SIMILAR PROTEIN 3} (\textit{CLPS3}, AT3G04680) is the closest ortholog of Potri.017G077200, and is related to the human Cleavage factor polyribonucleotide kinase subunit 1 (hCLP1), which forms part of the complex responsible for polyadenylation of 3' (3 prime) of messenger RNA~\cite{hCLP_de2000human}. CLPS3 also interacts with components of the polyadenylation complex in plants and it is expressed throughout whole plant development, including leaves and vasculature~\cite{CLP3_1}. Overexpression of \textit{CLPS3} causes aberrant leaf phenotypes, abnormal phyllotaxis, and early flowering~\cite{CLP3_1}. Futher, \textit{CLPS3} increases the expression of \textit{CUP-SHAPED COTYLEDON 1} (\textit{CUC1}), an NAC transcription factor, which together with \textit{CUC2} and \textit{CUC3}, have been found to participate in meristem formation, organ boundary separation, and leaf shape~\cite{cuc_postemb_hibara2006arabidopsis, cuc1_meristem_spinelli2011mechanistic, cuc2_leaf_nikovics2006balance}. Since Potri.017G077200 is an \textit{CLPS3} ortholog, it could play similar roles in leaf development of \textit{P. trichocarpa}, making it a strong target for genomic selection studies.
 
Potri.006G227300 is expressed in most plant tissues, but it is highly expressed in apical bud in spring, swelling bud, late dormant bud, as well as young and immature leaves~\cite{sreedasyam2022jgi}. Its \textit{Arabidopsis} ortholog, \textit{RHOMBOID-LIKE PROTEIN 12} (\textit{RBL12}, AT1G18600), follows a similar expression pattern, being enriched in floral buds~\cite{klepikova2016high}. Very little is know about \textit{RBL12}, but it is predicted to be an active transmembrane protease located in the mitochondria~\cite{RBL12_2_lemberg2007functional}. Further, \textit{RBL12} substrates in \textit{A. thaliana} have not been identified, therefore its role is yet to be determined~\cite{RBL12_1_kmiec2008plant}. Other genes that were associated with vein density by GWAS may play an indirect role in leaf development. For instance, the Potri.017G077300 ortholog, \textit{PBS1-LIKE 35} (\textit{PBL35}, AT3G01300), participates in shoot apical meristem homeostasis and plant immunity, while the Potri.006G090600 ortholog, \textit{ALDO-KETO REDUCTASE FAMILY 4 MEMBER C11} (\textit{AKR4C11}, AT3G53880), participates in abiotic stress tolerance through detoxification of reactive carbonyls~\cite{ark_rc_simpson2009characterization, ark_review_sengupta2015plant, pbl35_immunity_luo2020tyrosine, pbl35_sam_wang2022receptor}. Thus, Potri.017G077300 and Potri.006G090600 may play a role in leaf and vein development through such processes. Further, the Potri.006G227400 ortholog, \textit{CELL WALL INVERTASE 5} (\textit{CWINV5}, AT3G13784), is a cell wall invertase and members of this family have been found to affect plant development by making hexoses available for transport~\cite{cwi_1_sherson2003roles, klepikova2016high}.

\subsection{Conclusions}

Few-shot segmentation methods were extended to image-based plant phenotyping, whereby researchers can maximize predictive accuracy while minimizing the amount of training data. These methods were demonstrated for leaf scans of \textit{P. trichocarpa}, where 50 training images were used to train an automated tracing algorithm for whole-leaf segmentation and eight images were used to train a region growing algorithm to segment the visible vein architecture. {The methods were compared against a variant of the U-Net model, which suggested that the models considered here may produce more biologically realistic segmentations (i.e., reduce fragmentation).} The leaf and vein segmentations were used to extract biologically relevant morphological and topological traits related to the leaf body, venation, and petiole, which were validated with real-world manual measurements. Broad-sense clonal heritability estimates for each trait were measured, and a population-scale genomic analysis was conducted for vein density, which combined information from both leaf and vein segmentations. The GWAS analysis revealed a set of previously unconsidered SNPs and associated genes with mechanistic associations to multiple physiological processes relating to leaf development and function. Future work will include a deep dive into the relevant biology surrounding the features discussed in this work, and will include the extraction of additional cryptic phenotupes relating to leaf morphology and vein topology. In particular, this work will leverage systems biology, network analysis, and climatic data to uncover the mechanistic associations within and across genotypes as they relate to sustainable bioenergy applications (e.g., biomass yield and composition).

In conclusion, this study demonstrated a complete workflow from image acquisition to phenotype extraction. The utility of these methods for biological use cases was further demonstrated by {comparing biological accuracy with U-Net-based segmentations, and} performing GWAS which identified genomic regions and associated genes potentially controlling important plant phenotypes, such as vein density. This enhances current understanding of the genetic architecture of complex traits and may facilitate future quantitative genetics and genotype $\times$ environment interaction studies. This further allows researchers to assess how vein traits relate to other physiological processes, such as stomatal conductance, gas exchange, and overall plant productivity with important implications for developing \textit{Populus} as a bioenergy crop. Genes detected from the quantitative genetic analysis can be used in future biotechnology experiments for optimizing traits targeted for climate resilience, biomass production, and accelerated domestication for agriculture and biofuel production.

\section*{Acknowledgments}

\subsection*{Author Contributions} 

\noindent J. Lagergren: Conceptualization, funding acquisition, data collection, segmentation, baseline comparison, feature extraction, writing. \\
\noindent M. Pavicic: Conceptualization, data collection, feature extraction, writing. \\
\noindent H. Chhetri: Conceptualization, data collection, genomic analysis, writing. \\
\noindent L. York: Conceptualization, feature extraction, writing. \\
\noindent D. Hyatt: Genomic analysis, writing. \\
\noindent D. Kainer: Genomic analysis, writing. \\
\noindent E. Rutter: Segmentation, writing. \\
\noindent K. Flores: Segmentation, writing. \\
\noindent J. Bailey-Bale: Field site support, writing. \\
\noindent M. Klein: Field site support, writing. \\
\noindent G. Taylor: Field site support, writing. \\
\noindent D. Jacobson: Conceptualization, funding acquisition, supervision, writing. \\
\noindent J. Streich: Conceptualization, funding acquisition, data collection, writing. 

\subsection*{Special Thanks}

The authors would like to acknowledge members of the Taylor Lab (University of California, Davis): Zi (Janna) Meng, and Aiwei Zhu, for their support during data collection.

\subsection*{Funding}

This research used resources of the Oak Ridge Leadership Computing Facility, which is a DOE Office of Science User Facility supported under Contract DE-AC05-00OR22725. This work was funded by the Artificial Intelligence (AI) Initiative, an ORNL Laboratory Directed Research and Development program, and {by the Center for Bioenergy Innovation (CBI), which is a U.S. Department of Energy Bioenergy Research Center supported by the Office of Biological and Environmental Research in the DOE Office of Science. The sequencing work was conducted by the U.S. Department of Energy Joint Genome Institute, a DOE Office of Science User Facility, is supported by the Office of Science of the U.S. Department of Energy operated under Contract No. DE-AC02-05CH11231.} The manuscript was authored by UT-Battelle, LLC under Contract No. DE-AC05-00OR22725 with the US Department of Energy. The US Government retains and the publisher, by accepting the article for publication, acknowledges that the US Government retains a nonexclusive, paid-up, irrevocable, worldwide license to publish or reproduce the published form of this manuscript, or allow others to do so, for US Government purposes. The Department of Energy will provide public access to these results of federally sponsored research in accordance with the DOE Public Access Plan (\url{http://energy.gov/downloads/doe-public-access-plan}).

\subsection*{Conflicts of Interest}

The authors declare that they have no competing interests.

\subsection*{Data Availability}

In addition to releasing all of the segmentation code on a public GitHub repository~\cite{fsl_code}, we are also releasing all of the images, manual segmentations, model predictions, 68 extracted leaf phenotypes, and a new set of SNPs called against the v4 \textit{P. trichocarpa} genome for 1,419 genotypes on the Oak Ridge National Laboratory Constellation Portal (a public DOI data server)~\cite{fsl_data}. This is, to our knowledge, one of the largest releases of plant genotype and phenotype data in a single manuscript. We hope that this work becomes a valuable community resource and helps reduce barriers commonly associated with high throughput image-based plant phenotyping and machine learning.

\newpage



\section*{Supplementary Materials} \beginsupplement

\noindent \textbf{Video S1:} \textbf{Leaf segmentation video.} Animation of the leaf tracing algorithm, in which a CNN iteratively traces the boundary of a leaf. \underline{Left}: the raw leaf scan with an overlay of the previously traced path and a bounding box indicating the current position of the CNN model. \underline{Top right}: the image tile and with an overlay of the previously traced path that are input to the CNN. \underline{Bottom right}: the predicted pixels along the contour of the leaf that are used to update the position in the next iteration. The iteration proceeds until the CNN predictions reach the start of the trace. Note that in practice the iteration completes in $\sim$1 second, but is slowed down for better visualization.
\\

\noindent \textbf{Video S2:} \textbf{Vein segmentation video.} Animation of the vein growing algorithm, in which a CNN iteratively adds pixels to a growing segmentation of the visible vein architecture. \underline{Left}: the original leaf scan with an overlay of the pixels being considered by the CNN in yellow and classified vein pixels in red. \underline{Top right}: a zoomed in view of the top of the leaf and overlay. \underline{Bottom right}: a zoomed in view of the bottom of the leaf and overlay. The iteration proceeds, continuously adding new pixels to the segmentation, until no new pixels remain in the sample set. Note that in practice the iteration completes in $\sim$60 seconds, but is sped up for better visualization.

\newpage

\begin{figure}[ht!]
    \centering
    \includegraphics[width=\textwidth]{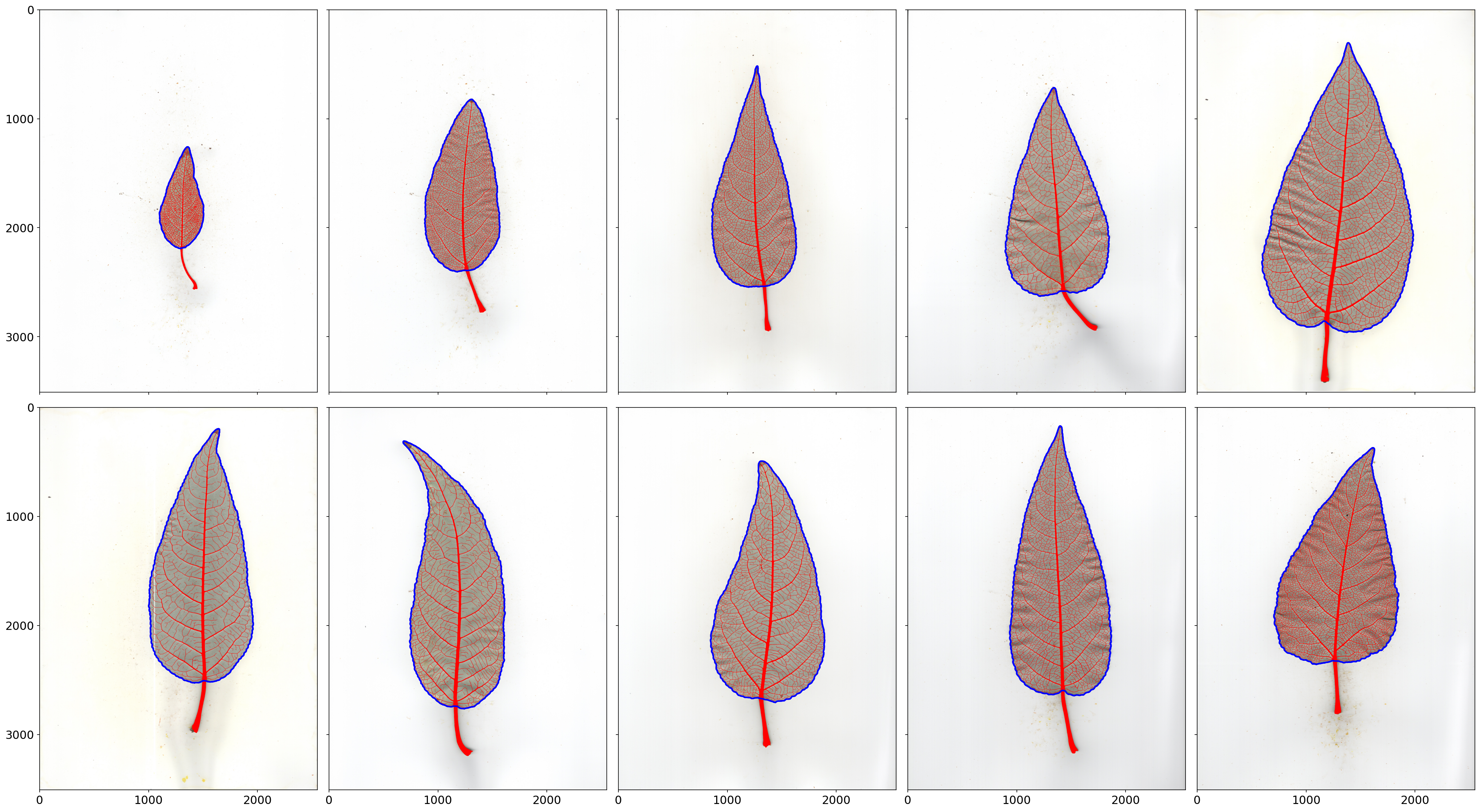}
    \caption{\textbf{Example leaf and vein segmentations.} Results of the leaf and vein segmentation methods on example leaf images outside the training set. Traced leaf contours are shown in blue and vein segmentations in red. \underline{Top}: segmentation overlays for leaves varying in size, going from smallest (left) to largest (right). \underline{Bottom}: segmentation overlays for leaves of approximately equal area, but varying in vein density, going from sparse (left) to dense (right) venation.}
    \label{fig:overlays}
\end{figure}

\newpage

\begin{figure}[ht!]
    \centering
    \includegraphics[width=\textwidth]{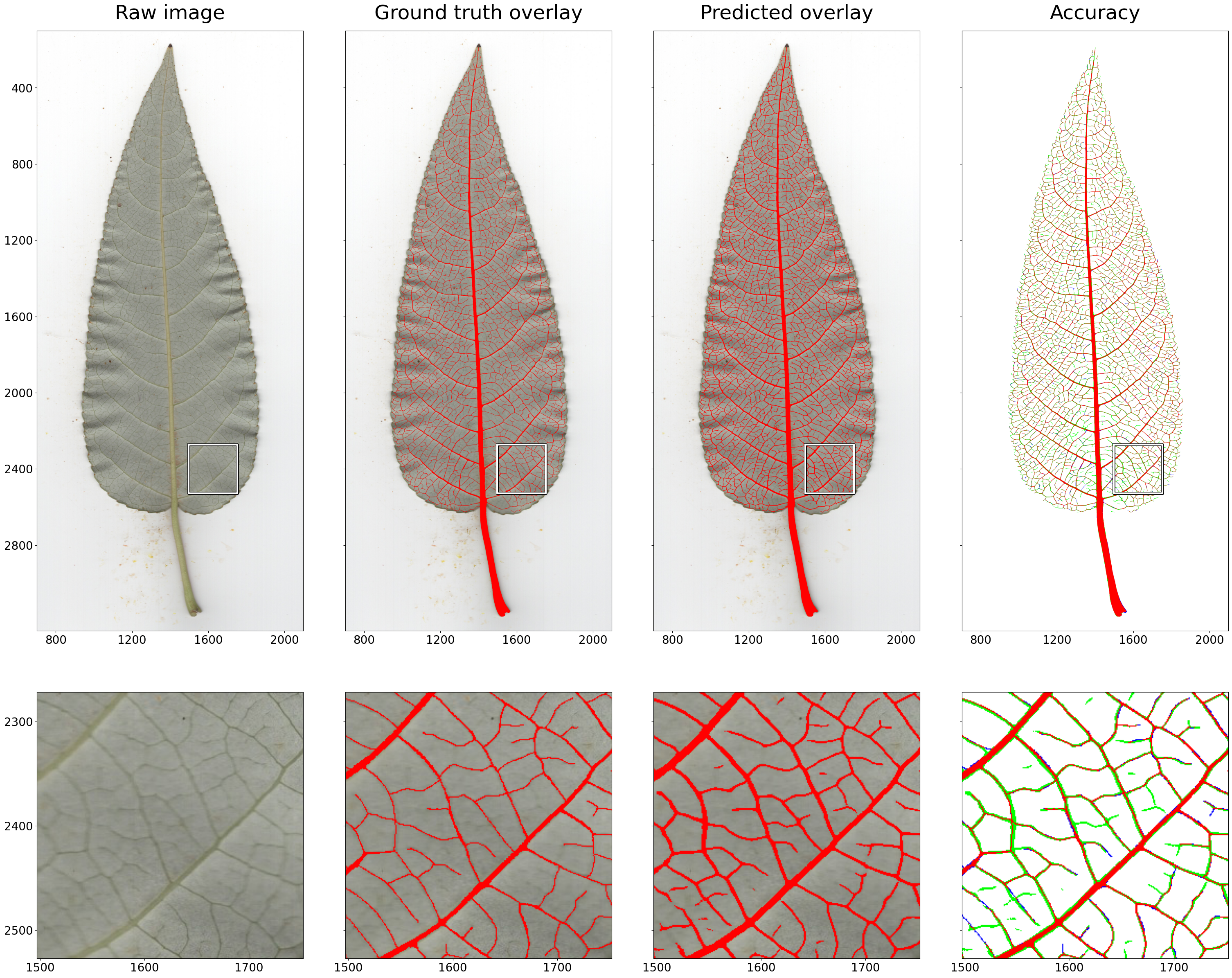}
    \caption{\textbf{Vein segmentation accuracy.} Results of the vein segmentation method on a leaf from the validation set. The top row shows the full leaf and the bottom row gives a zoomed in view. \underline{Left}: example leaf scan chosen from the validation set. \underline{Center left}: hand-annotated vein segmentation overlaid in red. \underline{Center right}: predicted vein segmentation overlaid in red. \underline{Right}: a comparison between the ground truth and predicted segmentations, in which red pixels indicate true positives, green pixels indicate false positives, and blue pixels indicate false negatives. Note that the zoomed-in tile reveals veins identified by the region growing method that are incorrectly reported as false positives (see veins with only green pixels) due to errors in the ground truth segmentation.}
    \label{fig:accuracy}
\end{figure}

\newpage


\newpage

\begin{longtable}[h!]{lcccp{7.5cm}}
    \caption{\textbf{Leaf features.} Includes names, units, broad-sense clonal heritability estimates, and descriptions of the 23 traits related to leaf morphology and color. Abbreviations: avg: average, max: maximum, min: minimum.} \label{tab:leaftraits} \\
    \hline
    \textbf{Feature} & \textbf{Units} & \textbf{$H^2$} & \textbf{Tool} & \textbf{Description} \\
    \hline
    \endfirsthead
    \textbf{Feature} & \textbf{Units} & \textbf{$H^2$} & \textbf{Tool} & \textbf{Description} \\
    \hline
    \endhead
    Area              & cm$^2$ & $0.30$ & Fiji & Total pixel count of leaf segmentation \\
    Aspect ratio      & -      & $0.58$ & Fiji & Ellipse major axis / ellipse minor axis \\
    Bottom blue       & -      & $0.57$ & Fiji & Avg. blue value of leaf abaxial side \\
    Bottom brightness & -      & $0.42$ & Fiji & Avg. brightness value of leaf abaxial side \\
    Bottom green      & -      & $0.41$ & Fiji & Avg. green value of leaf abaxial side \\
    Bottom hue        & -      & $0.39$ & Fiji & Avg. hue value of leaf abaxial side \\
    Bottom red        & -      & $0.45$ & Fiji & Avg. red value of leaf abaxial side \\
    Bottom saturation & -      & $0.27$ & Fiji & Avg. saturation value of leaf abaxial side \\
    Circularity       & -      & $0.23$ & Fiji & $4\pi A/P^2$ where $A$: area and $P$: perimeter \\
    Convex area       & mm$^2$ & $0.29$ & RVE  & Total pixel count of convex hull \\
    Major axis length & cm     & $0.21$ & Fiji & Major axis length of best-fit ellipse \\
    Minor axis length & cm     & $0.44$ & Fiji & Minor axis length of best-fit ellipse \\
    Max. Feret        & cm     & $0.23$ & Fiji & Max. distance between any two points in the leaf segmentation \\
    Min. Feret        & cm     & $0.42$ & Fiji & Min. distance between two parallel lines tangent to Max. Feret line \\
    Perimeter         & cm     & $0.25$ & Fiji & Sum of Euclidean distances between contour pixels in the leaf segmentation \\
    Roundness         & -      & $0.56$ & Fiji & $4A/(\pi M^2)$ where $A$: area, $M$: major axis \\
    Solidity          & -      & $0.09$ & Fiji & $A/C$ where $A$: area and $C$: convex area \\
    Top blue          & -      & $0.26$ & Fiji & Avg. blue value of leaf adaxial side \\
    Top brightness    & -      & $0.26$ & Fiji & Avg. brightness value of leaf adaxial side \\
    Top green         & -      & $0.23$ & Fiji & Avg. green value of leaf adaxial side \\
    Top hue           & -      & $0.29$ & Fiji & Avg. hue value of leaf adaxial side \\
    Top red           & -      & $0.24$ & Fiji & Avg. red value of leaf adaxial side \\
    Top saturation    & -      & $0.21$ & Fiji & Avg. saturation value of leaf adaxial side \\
    \hline      
\end{longtable}

\newpage


\begin{longtable}[h!]{lcccp{7.5cm}}
    \caption{\textbf{Vein features} Includes names, units, broad-sense clonal heritability estimates, and descriptions of the 27 traits related to vein morphology. Abbreviations: avg: average, DR: diameter range, max: maximum, min: minimum, RVE: RhizoVision Explorer.} \label{tab:veintraits} \\
    \hline
    \textbf{Feature} & \textbf{Units} & \textbf{$H^2$} & \textbf{Tool} & \textbf{Description} \\
    \hline
    \endfirsthead
    \textbf{Feature} & \textbf{Units} & \textbf{$H^2$} & \textbf{Tool} & \textbf{Description} \\
    \hline
    \endhead
    Area                 & mm$^2$ & $0.43$ & RVE    & Total pixel count of vein segmentation \\
    Area DR 1            & mm$^2$ & $0.55$ & RVE    & Projected area of veins with DR 0 - 0.25 mm \\
    Area DR 2            & mm$^2$ & $0.38$ & RVE    & Projected area of veins with DR 0.25 - 0.8 mm \\
    Area DR 3            & mm$^2$ & $0.26$ & RVE    & Projected area of veins with DR above 0.8 mm \\
    Avg. diameter        & mm     & $0.34$ & RVE    & Avg. skeletal pixel radius, doubled for diameter \\
    Convex area          & mm$^2$ & $0.29$ & RVE    & Total pixel count of convex hull \\
    Density              & -      & $0.65$ & Custom & Ratio of vein area to leaf area \\
    Length-to-area ratio & -      & $0.62$ & RVE    & $V/A$ where $V$: total length, $A$: leaf area \\
    Max. depth           & mm     & $0.24$ & RVE    & Max. vertical distance in vein segmentation \\
    Max. diameter        & mm     & $0.32$ & RVE    & Max. skeletal pixel radius, doubled for diameter \\
    Max. width           & mm     & $0.41$ & RVE    & Max. horizontal distance in vein segmentation \\
    Network solidity     & -      & $0.64$ & RVE    & Network Area per Convex Area ratio \\
    Perimeter            & mm     & $0.52$ & RVE    & Sum of Euclidean distances between contour pixels in the vein segmentation \\
    Surface area         & mm$^2$ & $0.46$ & RVE    & Length multiplied by cross-section circumference summed over skeletal pixels. \\
    Surface area DR 1    & mm$^2$ & $0.55$ & RVE    & Surface area of veins with DR 0 - 0.25 mm \\
    Surface area DR 2    & mm$^2$ & $0.38$ & RVE    & Surface area of veins with DR 0.25 - 0.8 mm \\
    Surface area DR 3    & mm$^2$ & $0.26$ & RVE    & Surface area of veins with DR above 0.8 mm \\
    Third order fraction & -      & $0.29$ & RVE    & Ratio of total length of DR 3 to total length \\
    Total length         & mm     & $0.53$ & RVE    & Sum of Euclidean distances between connected skeletal pixels \\
    Total length DR 1    & mm     & $0.56$ & RVE    & Total length of veins with DR 0 - 0.25 mm \\
    Total length DR 2    & mm     & $0.40$ & RVE    & Total length of veins with DR 0.25 - 0.8 mm \\
    Total length DR 3    & mm     & $0.27$ & RVE    & Total length of veins with DR above 0.8 mm \\
    Volume               & mm$^3$ & $0.29$ & RVE    & Length multiplied by cross-section area summed over skeletal pixels \\
    Volume DR 1          & mm$^3$ & $0.55$ & RVE    & Volume of veins with DR of 0 - 0.25 mm \\
    Volume DR 2          & mm$^3$ & $0.37$ & RVE    & Volume of veins with DR of 0.25 - 0.8 mm \\
    Volume DR 3          & mm$^3$ & $0.27$ & RVE    & Volume of veins with DR of above 0.8 mm \\
    Width-to-depth ratio & -      & $0.55$ & RVE    & Ratio of max. width to depth \\
    \hline
\end{longtable}

\newpage

\begin{longtable}[h!]{lcccp{7.5cm}}
    \caption{\textbf{Petiole features} Includes names, units, broad-sense clonal heritability estimates, and descriptions of the 18 traits related to petiole morphology and color. Abbreviations: avg: average, max: maximum, min: minimum. Note that Max. Feret is equivalent to petiole diameter that is used for validation in this work with real-world measurements.} \label{tab:petioletraits} \\
    \hline
    \textbf{Feature} & \textbf{Units} & \textbf{$H^2$} & \textbf{Tool} & \textbf{Description} \\
    \hline
    \endfirsthead
    \textbf{Feature} & \textbf{Units} & \textbf{$H^2$} & \textbf{Tool} & \textbf{Description} \\
    \hline
    \endhead
    Area              & cm$^2$ & $0.49$ & Fiji   & Total pixel count of petiole segmentation \\
    Aspect ratio      & -      & $0.41$ & Fiji   & Ellipse major axis / Ellipse minor axis \\
    Bottom blue       & -      & $0.29$ & Fiji   & Avg. blue value of petiole abaxial side \\
    Bottom brightness & -      & $0.25$ & Fiji   & Avg. brightness value of petiole abaxial side \\
    Bottom green      & -      & $0.27$ & Fiji   & Avg. green value of petiole abaxial side \\
    Bottom hue        & -      & $0.15$ & Fiji   & Avg. hue value of petiole abaxial side \\
    Bottom red        & -      & $0.22$ & Fiji   & Avg. red value of petiole abaxial side \\
    Bottom saturation & -      & $0.33$ & Fiji   & Avg. saturation value of petiole abaxial side \\
    Circularity       & -      & $0.45$ & Fiji   & $4\pi A/P^2$ where $A$: area and $P$: perimeter \\
    Major axis length & cm     & $0.52$ & Fiji   & Major axis length of the best-fit ellipse \\
    Minor axis length & cm     & $0.20$ & Fiji   & Minor axis length of the best-fit ellipse \\
    Max. Feret        & cm     & $0.55$ & Fiji   & Max. distance between any two points in the petiole segmentation \\
    Min. Feret        & cm     & $0.09$ & Fiji   & Min. distance between two parallel lines tangent to Max. Feret line \\
    Perimeter         & cm     & $0.55$ & Fiji   & Sum of Euclidean distances between contour pixels in the petiole segmentation \\
    Roundness         & -      & $0.39$ & Fiji   & $4A/(\pi M^2)$ where $A$: area, $M$: major axis \\
    Solidity          & -      & $0.10$ & Fiji   & $A/C$ where $A$: area and $C$: convex area \\
    Volume            & mm$^3$ & $0.43$ & RVE    & Length multiplied by cross-section area estimated from petiole diameter \\
    Width             & cm     & $0.25$ & Custom & Avg. diameter of the center 20\% of the petiole \\
    \hline
\end{longtable}

\newpage

\printbibliography

\end{document}